\documentclass[lettersize,journal]{IEEEtran}
\pdfoutput=1
\usepackage{amsmath,amsfonts}
\usepackage[caption=false,font=normalsize,labelfont=sf,textfont=sf]{subfig}
\usepackage{multirow}
\usepackage{bbding}
\usepackage{overpic}
\usepackage{bbm}
\usepackage{xcolor}
\usepackage{url}
\usepackage[colorlinks,linkcolor=red,anchorcolor=blue,citecolor=green]{hyperref}
\usepackage[super]{nth}

\definecolor{demphcolor}{RGB}{135,206,250}
\newcommand{\demph}[1]{\textcolor{demphcolor}{#1}}
\definecolor{demphgray}{RGB}{144,144,144}
\newcommand{\dempg}[1]{\textcolor{demphgray}{#1}}

\makeatletter

\newcommand{\Rmnum}[1]{\expandafter\@slowromancap\romannumeral #1@}
\makeatother

\begin{document}

\title{Attack-Augmentation Mixing-Contrastive Skeletal Representation Learning}

\author{Binqian~Xu, Xiangbo~Shu,~\IEEEmembership{Senior Member, IEEE}, Jiachao Zhang, Rui~Yan, and Guo-Sen Xie
\thanks{\textit{B. Xu, X. Shu, and G.-S. Xie are with the School of Computer Science and Engineering, Nanjing
			University of Science and Technology, Nanjing 210094, China. E-mail: \{xubinq11, gsxiehm\}@gmail.com, \{shuxb, jinhuitang\}@njust.edu.cn. Corresponding author: Xiangbo Shu.}}
   \thanks{\textit{J. Zhang is with the Artificial Intelligence Industrial Technology Research
Institute, Nanjing Institute of Technology, Nanjing 211167, China. E-mail:
zhangjc07@foxmail.com.}}
\thanks{\textit{R. Yan is with the School of Computer Science,
Nanjing University, Nanjing 210023, China. E-mail: ruiyan@nju.edu.cn.}}}

\markboth{Submission~of~IEEE~TRANSACTIONS~ON~PATTERN~ANALYSIS~AND~MACHINE~INTELLIGENCE, 2024}%
{Submission~of~IEEE~TRANSACTIONS~ON~PATTERN~ANALYSIS~AND~MACHINE~INTELLIGENCE, 2024}


\maketitle

\begin{abstract}
Contrastive learning, relying on effective positive and negative sample pairs, is beneficial to learn informative skeleton representations in unsupervised skeleton-based action recognition. To achieve these positive and negative pairs, existing weak/strong data augmentation methods have to randomly change the appearance of skeletons for indirectly pursuing semantic perturbations. However, such approaches have two limitations: i) solely perturbing appearance cannot well capture the intrinsic semantic information of skeletons, and ii) randomly perturbation may change the original positive/negative pairs to soft positive/negative ones. To address the above dilemma, we start the first attempt to explore an attack-based augmentation scheme that additionally brings in direct semantic perturbation, for constructing hard positive pairs and further assisting in constructing hard negative pairs. In particular, we propose a novel Attack-Augmentation Mixing-Contrastive skeletal representation learning (A$^2$MC) to contrast hard positive features and hard negative features for learning more robust skeleton representations. In A$^2$MC, Attack-Augmentation (Att-Aug) is designed to collaboratively perform targeted and untargeted perturbations of skeletons via attack and augmentation respectively, for generating high-quality hard positive features. Meanwhile, Positive-Negative Mixer (PNM) is presented to mix hard positive features and negative features for generating hard negative features, which are adopted for updating the mixed memory banks. Extensive experiments on three public datasets demonstrate that A$^2$MC is competitive with the state-of-the-art methods. The code will be accessible on \href{https://github.com/1xbq1/A2MC}{A$^2$MC}.
\end{abstract}


\section{Introduction}
\label{sec:intro}

\IEEEPARstart{H}{uman} action recognition is a fundamental topic in the field of computer vision with various application scenarios, e.g., video surveillance, human-machine interaction, entertainment, etc.~\cite{carreira2017quo,shu2019hierarchical,shu2020host,kumawat2022action,li2022egocentric}. Due to the lightweight and robustness of skeletons, skeleton-based action recognition has attracted a lot of attention~\cite{kim2017interpretable,li2018independently,shu2021spatiotemporal,shu2022multi}. In recent years, supervised methods have achieved satisfactory performance, but the requirement for sufficient labeled data remains an obstacle~\cite{li2019actional,wang2021iip,yan2018spatial}. Alternatively, many unsupervised methods in the field have emerged endlessly, mainly based on the following three types of models, i.e., encoder-decoder~\cite{su2020predict,zheng2018unsupervised}, contrastive learning~\cite{guo2022contrastive,li20213d}, and Transformer~\cite{cheng2021hierarchical,kim2022global}. Among them, contrastive learning-based methods are naturally specific for unsupervised tasks, and continue to be attractive in the community~\cite{chen2022hierarchically,guo2022contrastive,kim2022global}.

Generally, the process of contrastive learning is to pull the positive pairs closer and push the negative pairs away for learning representation of data~\cite{chen2020simple,he2020momentum}. Therefore, the key to contrastive learning is to construct positive (similar) pairs and negative (dissimilar) pairs~\cite{gutmann2010noise,robinson2020contrastive}. In the unsupervised skeleton-based action recognition task, weak/strong augmentation is the most general way to construct positive and negative features~\cite{guo2022contrastive,li20213d,lin2020ms2l}, which always performs random appearance perturbation for indirectly realizing the semantic perturbation. 
\begin{figure}[!t]
\begin{center}
\includegraphics[width=1.0\linewidth]{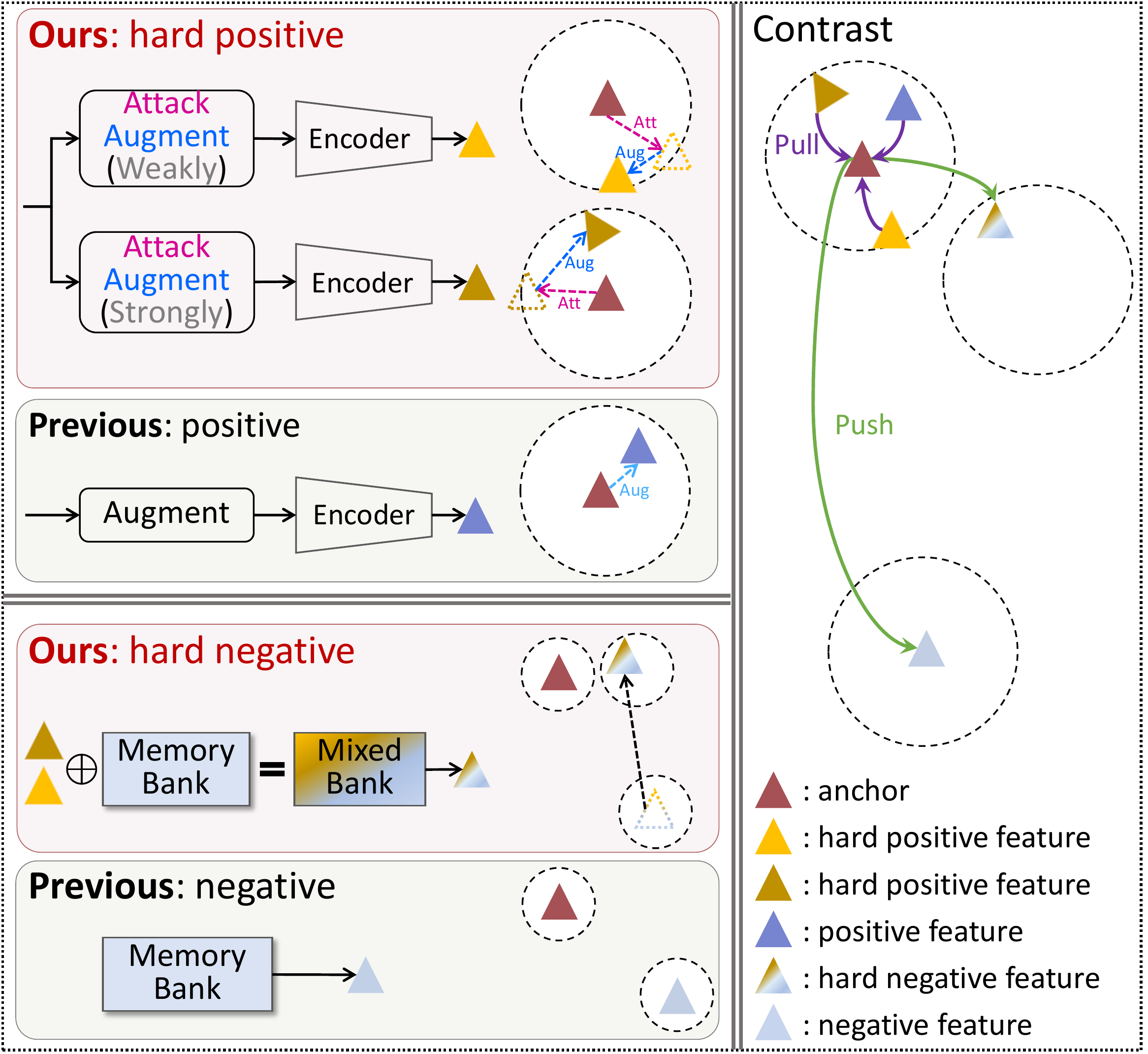}
\end{center}
   \caption{Main idea of this work. Previous weak/strong augmentations randomly construct positive features for the anchor while directly treating them as negative features in the memory bank. Our work introduces an attack-augmentation ({\bf ``Att": attack features for moving them toward the semantic boundary, and ``Aug": randomly augment features for diversity}) to generate hard positive features, and further construct positive-negative mixed features as hard negative features. Compared with positive/negative features, our hard positive/negative features are more beneficial for contrastive learning.}
\label{fig:idea}

\end{figure}
However, solely perturbing the appearance of skeletons cannot well capture the intrinsic semantic information of skeletons, due to that the indirect semantic perturbation caused by appearance perturbation is not consistent with direct semantic perturbation. Besides, randomly perturbing appearance may change the original positive/negative pairs to soft positive/negative ones. Compared with soft positive/negative features, hard positive/negative features are more beneficial for contrastive learning.  Intuitively, a desirable hard positive feature should be close to the semantic inner boundary of the anchor, namely it should be far away from the anchor as far as possible.
Similarly, a desirable hard negative feature should be close to the semantic outer boundary of the anchor~\cite{kalantidis2020hard,robinson2020contrastive}. 

Thus, how to effectively move positive and negative features toward the semantic inner and outer boundaries of the anchor respectively is the key to constructing desirable hard positive pairs and hard negative pairs. The semantic boundary refers to the border between action categories. Since the original category is unknown, flattening the confidence of features across all categories to a certain extent decreases the confidence in the original category, making features closer to the semantic boundary.
The above idea is inspired by the related method of a white-box attack~\cite{liu2020adversarial,wang2021understanding}. Based on this, we start the first attempt to design an Attack-Augmentation (Att-Aug) mechanism equipped with weak/strong attack-augmentation to realize the direct semantic perturbation and appearance perturbation, for effectively generating hard positive features, and further assisting in constructing hard negative features. As shown in~\ref{fig:idea}, weak/strong attack-augmentation performs the semantic perturbation of skeleton by attacking the feature for being close to the semantic boundary, as well as performing the appearance perturbation for producing more diverse hard positive features. Subsequently, these hard positive features are further mixed with negative features from an updating memory bank to generate hard negative features, which are adopted for updating the mixed memory banks.

\begin{figure*}[!t]
\begin{center}
\begin{overpic}[width=0.97\linewidth]{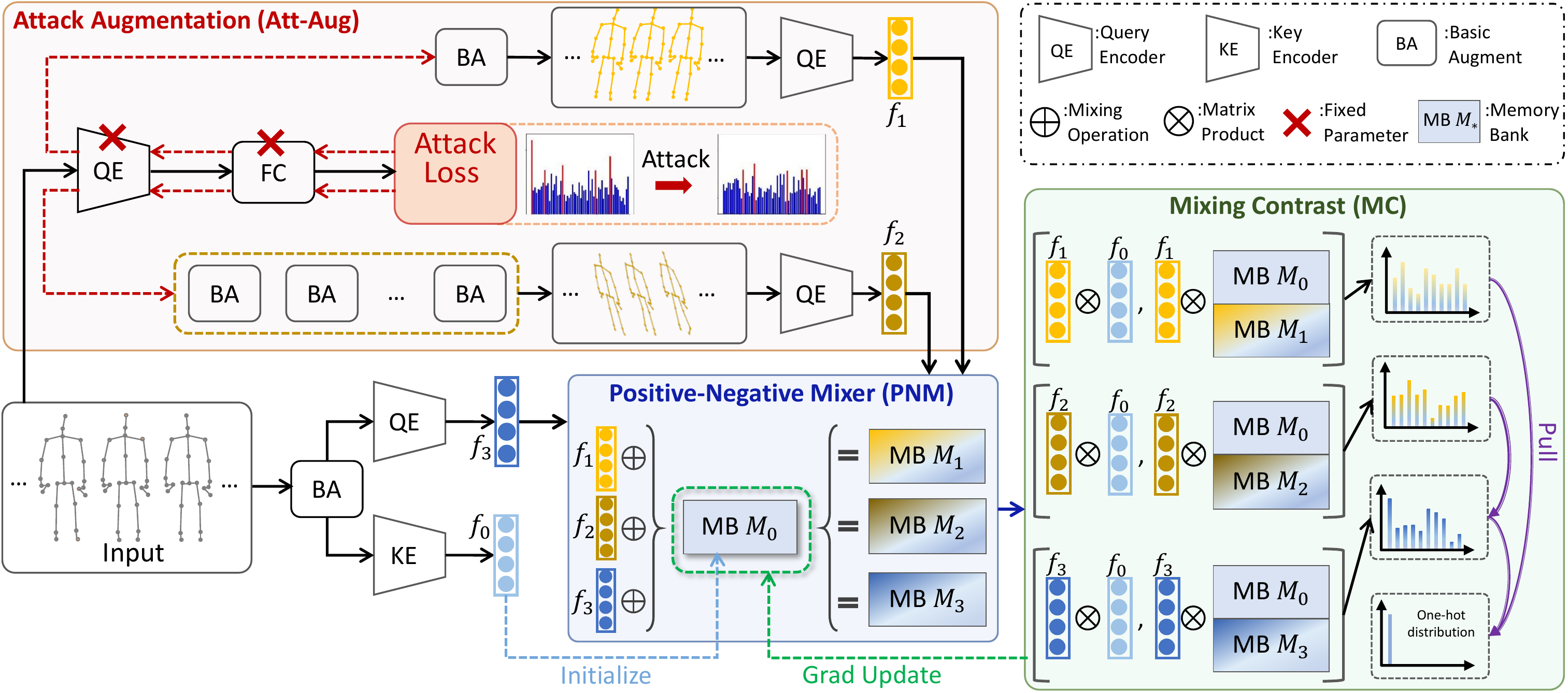}
\put(26.2,31){\tiny (Eq.~\ref{eq:att1}-\ref{eq:att2})}
\put(60.5,0.69){\tiny (Eq.~\ref{eq:M0})}
\end{overpic}
\end{center}
\vspace{-2mm}
   \caption{Framework of the proposed ${\text A}^2$MC. Given the skeleton input, Att-Aug produces two types of attacked features $f_1/f_2$ (weak and strong versions) as hard positive features (\S\ref{sec:att-aug}), while features $f_0/f_3$ are obtained via basic augment and key$/$query encoder (\S\ref{sec:con}). PNM mixes features $f_1/f_2/f_3$ and memory bank $M_0$ to generate mixed memory banks $M_1/M_2/M_3$, in which memory bank $M_0$ is an ensemble of adversarial negative features updated by gradient, initialized by features $f_0$ (\S\ref{sec:pnm}). In MC, the similarity distributions calculated from $f_1$ and $f_3$ are respectively pulled to the one-hot distribution, and the similarity distribution calculated from $f_2$ is pulled to the similarity distribution calculated from $f_3$ (\S\ref{sec:mc}).}
\label{fig:framework}

\end{figure*}

Based on Att-Aug, we propose a novel Attack-Augmentation Mixing-Contrastive skeletal representation learning (A$^2$MC) framework for unsupervised skeleton-based action recognition, as shown in~\ref{fig:framework}. Specifically, Attack-Augmentation (Att-Aug) mechanism is used to construct hard positive features, in which the skeletons are first updated by the attack loss to get close to the semantic boundary, and then passed through weak/strong augmentations and query encoders to obtain abundant hard positive features. Then, Positive-Negative Mixer (PNM) mixes a small proportion of hard positive features and a large proportion of adversarial negative features (in a learnable memory bank) to generate hard negative features (in the mixed memory banks). Finally, Mixing Contrast (MC) loss trains the entire network by pulling similarity distribution closer to one-hot distribution or similarity distribution calculated from basic augmented features. Experiments on NTU RGB+D 60, NTU RGB+D 120, and PKU-MMD demonstrate the effectiveness of the proposed A$^2$MC on the unsupervised skeleton-based action recognition task. Overall, the main contributions of this work can be summarized as follows,
\begin{itemize}
	\item We propose a novel Attack-Augmentation Mixing-Contrastive skeletal representation learning (${\text A}^2$MC) framework that effectively constructs hard positive and negative pairs, enabling the model to learn more robust skeleton representations for unsupervised skeleton-based action recognition.
        \item To explore richer hard positive features, we design a new Attack-Augmentation (Att-Aug) mechanism that attacks and augments skeleton sequences mainly for the targeted semantic perturbation and untargeted appearance perturbation, respectively.
        \item To construct and update hard negative features, we present a new Positive-Negative Mixer (PNM) that mixes hard positive features and adversarial negative features, and then updates the mixed memory banks.
\end{itemize}

\section{Related Work}
\label{sec:related}

\noindent\textbf{Unsupervised skeleton action recognition.}
Many unsupervised methods have emerged for skeleton-based action recognition that aim to learn representations from unlabeled data~\cite{chen2022hierarchically,guo2022contrastive,kim2022global,li2020sparse,li20213d,lin2020ms2l,mao2022cmd,si2020adversarial,su2020predict,su2021modeling,su2021self,thoker2021skeleton,tu2022joint,xu2021prototypical,xu2021unsupervised,yang2021skeleton,yao2021recurrent,zhang2022contrastive,mao2023masked}. Existing methods almost belong to three categories: encoder-decoder, contrastive learning, and Transformer. Since the emergence of MoCo~\cite{chen2020improved,he2020momentum} in 2020, contrastive learning-based methods have been extensively studied. For example, Inter-skeleton
Contrast (ISC)~\cite{thoker2021skeleton} is proposed to learn multiple different skeleton representations in an inter- and intra-contrastive manner. Beyond single-view contrastive learning, CrosSCLR~\cite{li20213d} additionally explores cross-view contrast between joint and motion views.
In ActCLR~\cite{lin2023actionlet}, the motion and static parts are augmented to different degrees to construct more diverse positive pairs for contrastive learning. 
Recently, Cross-modal Mutual Distillation (CMD)~\cite{mao2022cmd} utilizes the k-nearest neighbors of the anchor in the memory bank for constructing more positive pairs, and explores the neighboring similarity distribution in the contrastive frameworks.
In this work, the proposed A$^2$MC is also evolved from MoCo by using the mixing memory bank instead of the traditional memory bank to address the problem of unsupervised skeleton-based action recognition.

With the primary objective of generating hard positive pairs, HaLP~\cite{shah2023halp} minimizes the similarity between the generated positive pairs and real positive pairs while ensuring that both belong to the same prototype. A$^2$MC not only generates hard positive pairs via gradient-based white-box attack, but also creates hard negative pairs by mixing hard positive ones with negative features in the memory bank.
%


\noindent\textbf{Contrastive learning.}
Contrastive learning aims to learn data representation by measuring the similarity/dissimilarity on the constructed positive/negative pairs~\cite{chen2020simple,gutmann2010noise,he2020momentum,robinson2020contrastive}. Many methods consider different augmentation strategies to construct positive features for the anchor feature~\cite{chen2020simple,guo2022contrastive,lin2020ms2l,rao2021augmented,thoker2021skeleton,zhang2022hierarchical}. Among them, AS-CAL~\cite{rao2021augmented} utilizes seven different normal augmentations to produce the positive features, such as rotate, shear, reverse and so on. Besides the normal augmentations, AimCLR~\cite{guo2022contrastive} combines eight augmentation strategies to produce an extreme augmentation to enrich diverse features. To alleviate the inconsistency caused by extreme augmentation, Zhang et al.~\cite{zhang2022hierarchical} introduce a step-by-step hierarchical augmentation to construct more consistent positive pairs. Moreover, some studies prove that hard negative features are also important for learning robust representations on contrastive learning~\cite{kalantidis2020hard,robinson2020contrastive}. Likewise, the hard positive pairs may be also important, which is almost ignored in this community. In this work, we explore the characteristic of hard positive features, and aim to simultaneously construct hard positive and negative features.

\noindent\textbf{Gradient-based white-box attack.}
Deep learning models are still vulnerable to adversarial perturbations and attacks~\cite{szegedy2013intriguing, tanaka2022adversarial,zhang2024meta}, even though they have achieved outstanding performance. Therefore, various adversarial attacks have been explored~\cite{akhtar2021advances,goodfellow2014explaining,madry2017towards}. Among them, White-box Attack is one of the most representative attacks, wherein all knowledge of the target model is known~\cite{akhtar2021advances}. In the White-box attack family, the gradient-based attack is one of the mainstream technologies used to fool the model by ascending the gradient on the model loss surface~\cite{goodfellow2014explaining}. Recently, a few works~\cite{liu2020adversarial,wang2021understanding} attempt to leverage the gradient-based attack to test the robustness of the learning model. In this work, we explore the gradient-based White-box Attack to generate hard positive features for further improving the robustness of contrastive representation.

\section{Methodology}
\subsection{Preliminary}
\label{sec:con}
Given the skeleton sequence $x$, we first perform basic data augmentation~\cite{thoker2021skeleton} (including pose augmentation or joint jittering, and temporal crop-resize) on it to obtain two augmented versions $x_0$ and $x_3$, following MoCo v2~\cite{chen2020improved}. Subsequently, $x_0$ and $x_3$ are input into the key encoder and query encoder to produce features $f_0$ and $f_3$, respectively. A memory bank ${M}=\{m_i\}_{i=1}^K$ stores a lot of negative features from $f_0$ in a first-in-first-out way. InfoNCE loss~\cite{oord2018representation} is employed to train the whole contrastive network, i.e.,
\begin{equation}
\label{eq_infonce}
    \mathcal{L}_{in} = -\log\frac{\exp(f_3^\top f_0/\tau)}{\exp(f_3^\top f_0/\tau) + \sum\nolimits_{i=1}^K\exp(f_3^\top m_i/\tau)}
\end{equation}
where $\tau$ is the temperature hyper-parameter. Similar to~\cite{guo2022contrastive}, we can also convert the above InfoNCE loss into another version for later understanding. The first step is to calculate the similarity distribution of $u$ and $v$ as follows,
\begin{equation}
    \psi(u,v) = \frac{\exp(u^\top v/\tau)}{\exp(u^\top v/\tau) + \sum\nolimits_{i=1}^{K}\exp(u^\top m_i/\tau)}
\end{equation}
Thus, another version of InfoNCE loss is expressed as
\begin{equation}
\label{eq_infonce_v1}
    \mathcal{L}_{in} = - {\bf 1}\cdot \log\psi(f_3,f_0) - \sum\nolimits_{i=1}^K {\bf 0}\cdot \log\psi(f_3,m_i)
\end{equation}
where {\bf 1} and {\bf 0} are the element indicators in the one-hot distribution. To maintain the consistency of the features stored in the memory bank $M$, the key encoder is a momentum-updated version of the query encoder, as follows,
\begin{equation}
    \theta_k \leftarrow \alpha\theta_k + (1-\alpha)\theta_q
\end{equation}
where $\alpha$ is momentum coefficient, and $\theta_k$/$\theta_q$ represents the parameter sets of key/query encoder. 

\subsection{Attack-Augmentation (Att-Aug)}
\label{sec:att-aug}
Many contrastive-based methods adopt weak/strong augmentations to generate positive pairs~\cite{guo2022contrastive,li20213d,lin2020ms2l,mao2022cmd,rao2021augmented,zhang2022hierarchical}, but rarely explore how to generate real hard positive pairs. Intuitively, hard positive pairs should be mostly closer to the semantic inner boundary, which can increase contrasting capability for learning more robust representations. Thus, to construct hard positive features, we design a new Attack-Augmentation (Att-Aug) mechanism based on gradient-based White-box Attack, which collaboratively performs the targeted semantic attack and untargeted appearance augmentation. Specifically, the former aims to move the skeleton features to the semantic boundary of the anchor, and the latter randomly moves skeleton features for making them diverse.

\noindent\textbf{Targeted semantic attack.} One way to approach the semantic boundary of the anchor is to smooth its semantic distribution, to some extent, moving away from the original class semantics and closer to the semantics of other classes. It should be noted that the ``targeted" here refers to a targeted change of semantics, in contrast to the random semantic change in the following ``untargeted appearance augmentation".
Specifically, the skeleton sequence $x$ is first input into a fixed query encoder and a fully connected layer in turn to obtain its class feature $f_a\in {\mathcal{R}}^{ C\times1}$. Second,  
we formulate an attack loss to smooth $f_a$ for moving the distribution of $f_a$ towards a uniform distribution, so that the attacked $\hat{x}$ is relatively closer to the semantic boundary compared with the original itself (anchor). 
It is worth noting that pulling the distribution of $f_a$ toward a uniform distribution will definitely decrease the score of its assigned class while increasing the scores of the remaining classes to some extent. As a result, $f_a$ is brought closer to the boundary between its own class and the remaining classes.
Formally, the attack loss is formulated to maximize the entropy of $f_a$, as follows,
\begin{equation}
\label{eq:att1}
\begin{aligned}
    \mathcal{L}_{at}&=-\text{Entropy}(f_a) \\
    &\stackrel{\triangle}{=}\text{MSE}(f_a-\frac{1}{C}\mathbbm{1})
\end{aligned}
\end{equation}
where $\text{MSE}(~)$ denotes the mean square error function, realizing the function of maximizing entropy to facilitate derivation, and $\mathbbm{1}$ is an all-ones vector. 
After the attack loss is computed, the gradient of $x$ is $\frac{\partial \mathcal{L}_{at}}{\partial x}$ for updating the skeleton sequence. And the attacked skeleton sequence $\hat{x}$ is calculated as
    \begin{equation}
    \label{eq:att2}
    \begin{aligned}
        \rho &= \epsilon\Phi(\frac{\partial \mathcal{L}_{at}}{\partial x}, x) \\
        \hat{x} &= x + \rho,\quad\Vert \rho \Vert_2 < \eta
    \end{aligned}
    \end{equation}
where $\epsilon$ is the learning rate, $\Phi$ is Adam optimizer~\cite{kingma2014adam} to compute the updates, $\rho$ denotes the perturbation restricted by $\ell_2$-norm, and $\eta$ is a scalar value to ensure a certain imperceptibility in skeletons. 
So far, the skeleton sequence is closer to the semantic boundary of the anchor in the semantic space after the targeted semantic attack.

\noindent\textbf{Untargeted appearance augmentation.} Given the attacked skeleton sequence $\hat{x}$, weak and strong augmentations are performed on it to change its appearance, respectively. Specifically, the weak augmentation is consistent with the basic augmentation of contrastive learning, including pose augmentation or joint jittering, and temporal crop-resize. The strong augmentation consists of more basic augmentations~\cite{guo2022contrastive}: pose augmentation or joint jittering, temporal crop-resize, spatial flip, rotate, gaussian noise, gaussian filter, and axis mask. Subsequently, weak$/$strong augmented skeleton sequences are converted to hard positive features $f_1/f_2$ via query encoders.

\subsection{Positive-Negative Mixer (PNM)}
\label{sec:pnm}
It is learned from some studies that hard negative features can also increase contrasting capability~\cite{kalantidis2020hard,robinson2020contrastive}. In this work, Positive-Negative Mixer (PNM) aims to mix negative features from an online updating memory bank and positive features ($f_1/f_2/f_3$) according to different weights, and further generate hard negative features participating in the updating of the mixed memory bank, which is detailed in the following.


\noindent\textbf{Gradient-updating memory bank.} Different from the traditional memory bank of basic contrastive learning in a first-in-first-out partial updating way, the memory bank $M_0$ of PNM as a whole is updated in the direction closer to the anchor by the gradient~\cite{hu2021adco,qi2022adversarial}. First, $M_0$ is initialized by storing features from the key encoder. Suppose the total loss is calculated as $\mathcal{L}$ after the forward propagation, the gradient respect to $M_0$ can be denoted as $\frac{\partial \mathcal{L}}{\partial M_0}$. Subsequently, the updating process of $M_0$ is expressed as follows,
\begin{equation}
\label{eq:M0}
    M_0 \leftarrow M_0 + \beta\frac{\partial \mathcal{L}}{\partial M_0}
\end{equation}
where $\beta$ is the learning rate. 

\noindent\textbf{Positive-negative mixing.}
To further generate hard negative features based on $M_0$, the strategy of mixing positive and negative features online is adopted. Specifically, given the positive features ($f_1$, $f_2$, and $f_3$) and negative features from $M_0$, the mixing process for generating memory banks $M_1$, $M_2$, and $M_3$ is calculated as follows,
\begin{equation}
\label{eq:mix}
\begin{aligned}
    M_* &= \{m_{*i}\}_{i=1}^K = \{\lambda f_* + (1-\lambda) m_{0i}\}_{i=1}^K \\
\end{aligned}
\end{equation}
where the subscript $*\in\{1,2,3\}$, $K$ is the length of memory bank $M_*$, and $\lambda$ is a mixing coefficient. $\lambda$ is usually less than $0.5$ to ensure that the mixed features are closer to the negative features. Here, due to the import of the semantic of hard positive features, the generated hard negative features is relatively close to the semantic outer boundary of the anchor. 

\subsection{Mixing Contrast (MC)}
\label{sec:mc}
In this work, Mixing Contrast (MC) loss is leveraged to contrast all positive and negative features, which pulls the similarity distributions among positive and negative features closer to the corresponding distributions, as shown in~\ref{fig:framework}. Formally, given all types of positive features $\{f_0,f_1,f_2,f_3\}$, and memory banks $\{M_0,M_1,M_2,M_3\}$, MC loss $\mathcal{L}$ is defined as the final contrastive loss, i.e.,
\begin{equation}
\label{MCloss}
    \mathcal{L} = \mathcal{L}_{1} + \mathcal{L}_{2} + \mathcal{L}_{3}
\end{equation}
In $\mathcal{L}_{1}$, the similarity distribution of $f_0$, $f_1$, $M_0$, and $M_1$ is pulled toward the one-hot distribution. $M_0$ and $M_1$ are concatenated to generate the final memory bank $M_{01}$, and then $\mathcal{L}_{1}$ is calculated as follows,
\begin{equation}
\label{eq_infonce_final}
\begin{aligned}
    M_{01} &= M_0 \cup M_1 = \{m_i\}_{i=1}^{2K} \\
    \mathcal{L}_{1} &= - {\bf 1}\cdot \log\psi(f_1,f_0) - \sum\nolimits_{i=1}^{2K} {\bf 0}\cdot \log\psi(f_1,m_i)
\end{aligned}
\end{equation}
The loss formulas for $\mathcal{L}_{3}$ and $\mathcal{L}_{1}$ are identical, $\mathcal{L}_{1}$($\mathcal{L}_{3}$) maximizes the agreement of $f_1$($f_3$) and $f_0$($f_0$) while minimizing the agreement of $f_1$($f_3$) and $M_{01}$($M_{03}$).
Different from the above pulling of one-hot distribution, $\mathcal{L}_{2}$ needs to pull the similarity distribution of $f_0$, $f_2$, $M_0$, and $M_2$ toward the similarity distribution from basic augmentation, due to the strong augmentation~\cite{guo2022contrastive}, as follows,
\begin{equation}
\label{eq:l_02}
\begin{aligned}
    M_{02} = &M_0 \cup M_2 = \{m_i\}_{i=1}^{2K} \\
    \mathcal{L}_{2} = &- \psi(f_3,f_0)\cdot \log\psi(f_2,f_0) \\
    &- \sum\nolimits_{i=1}^{2K} \psi(f_3,m_i)\cdot \log\psi(f_2,m_i)
\end{aligned}
\end{equation}
Finally, MC loss $\mathcal{L}$ in Eq.~\eqref{MCloss} is used to train the whole framework of ${\text A}^2$MC.

\section{Experiments}
\subsection{Datasets}
\noindent\textbf{NTU RGB+D 60 (NTU-60)~\cite{shahroudy2016ntu}:} contains 56,880 skeleton sequences in 60 classes collected from 40 subjects by three cameras. In x-sub benchmark, training sets come from 20 subjects while testing sets come from the other 20 subjects. In x-view benchmark, training sets come from camera 2 and 3 while testing sets come from camera 1. 

\noindent\textbf{NTU RGB+D 120 (NTU-120)~\cite{liu2019ntu}:} contains 114,480 skeleton sequences in 120 classes collected from 106 subjects with 32 setups, extended from NTU RGB+D 60. In x-sub benchmark, training sets come from 53 subjects while testing sets come from the other 53 subjects. In x-set benchmark, training sets come from even-numbered setups while testing sets come from odd-numbered setups.

\noindent\textbf{PKU-MMD~\cite{liu2017pku}:} contains nearly 20,000 action clips in 51 classes, and consists of Phase \Rmnum{1} and Phase \Rmnum{2} (PKU-\Rmnum{2}) collected from Kinect cameras. PKU-\Rmnum{2} is more challenging due to the large view variation, which is recommended to validate the model. PKU-\Rmnum{2} contains 6,945 skeleton sequences in 41 action classes, performed by 13 subjects. In the recommended cross-subject benchmark, the training sets and testing sets include 5,332 and 1,613 skeleton sequences, respectively.

\subsection{Implementation Details and Evaluation}
Following the previous related works~\cite{mao2022cmd,thoker2021skeleton}, BiGRU is adopted as the encoder for a fair comparison. Except for the experiments in Table~\ref{li_compare}, which use different encoders such as Transformer~\cite{shi2020decoupled}, GCN~\cite{yan2018spatial}, and GCN \& Transformer~\cite{wu2024scd}, all other tables use BiGRU as the encoder. All sequence lengths are resized to the fixed 64 frames via temporal crop-resize~\cite{thoker2021skeleton}. We adopt SGD with momentum 0.9 and weight decay $10^{-4}$ to optimize the proposed ${\text A}^2$MC. The initial learning rate and the mini-batch size are set to 0.01 and 64, respectively. All experiments are performed on the PyTorch framework~\cite{paszke2017automatic}.

\renewcommand{\arraystretch}{1.3}
\begin{table*}[!t]
\caption{Linear evaluation results in different methods on NTU-60, NTU-120, and PKU-\Rmnum{2}. The best and second-best values are highlighted in {\textbf{bold}} and \underline{underlined}, respectively.}
\label{li_compare}
\centering    
  \setlength{\tabcolsep}{8pt}
\begin{tabular}{l|c|c|c|c|c|c|c|c}
\hline
\multirow{2}{*}{Method}&
\multirow{2}{*}{Encoder}&
\multirow{2}{*}{FLOPs} &
\multirow{2}{*}{Params}&
\multicolumn{2}{c|}{NTU-60}&
\multicolumn{2}{c|}{NTU-120}&
PKU-\Rmnum{1} \\
\cline{5-9}
&&&&x-sub & x-view & x-sub & x-set & x-sub \\ \hline\hline
MS$^2$L~\cite{lin2020ms2l}$_{\text{\demph{(ACM MM 20)}}}$ & GRU & - & 2.28M & 52.6 & - & - & -  & 64.9 \\
PCRP~\cite{xu2021prototypical}$_{\text{\demph{(IEEE TMM 21)}}}$ & GRU & - & - & 54.9 & 63.4 & 43.0 & 44.6 & - \\
LongT GAN~\cite{zheng2018unsupervised}$_{\text{\demph{(AAAI 18)}}}$ & GRU & - & 40.2M & 39.1 & 48.1 & - & - & - \\
EnGAN~\cite{kundu2019unsupervised}$_{\text{\demph{(WACV 19)}}}$ & LSTM & - & - & 68.6 & 77.8 & - & - & 85.9 \\
P$\&$C~\cite{su2020predict}$_{\text{\demph{(CVPR 20)}}}$ & GRU & - & - & 50.7 & 76.3 &  42.7 & 41.7 & - \\
SeBiReNet~\cite{nie2020unsupervised}$_{\text{\demph{(ECCV 20)}}}$ & GRU & - & 0.27M & - & 79.7 & - & - & - \\
TS Colorization~\cite{yang2021skeleton}$_{\text{\demph{(ICCV 21)}}}$ & GCN & - & - & 71.6 & 79.9 & - & - & - \\
H-Transformer~\cite{cheng2021hierarchical}$_{\text{\demph{(ICME 21)}}}$ & Transformer & & $>$100M & 69.3 & 72.8 & - & - & - \\
GL-Transformer~\cite{kim2022global}$_{\text{\demph{(ECCV 22)}}}$ & Transformer & 118.62G & 214M & 76.3 & 83.8 & 66.0 & 68.7 & - \\
AS-CAL~\cite{rao2021augmented}$_{\text{\demph{(INS 21)}}}$ & LSTM & - & 0.43M & 58.5 & 64.8 & 48.6 & 49.2 & - \\
ISC~\cite{thoker2021skeleton}$_{\text{\demph{(ACM MM 21)}}}$ & GRU\&CNN\&GCN & 5.76G & 10.0M & 76.3 & 85.2 & 67.1 & 67.9 & 80.9 \\
CrosSCLR~\cite{li20213d}$_{\text{\demph{(CVPR 21)}}}$ & GCN & 5.76G & 0.85M & 72.9 & 79.9 & - & - & -  \\
CrosSCLR-B~\cite{li20213d}$_{\text{\demph{(CVPR 21)}}}$ & BiGRU & 5.76G & 10.0M & 77.3 & 85.1 & 67.1 & 68.6 & - \\
CRRL~\cite{wang2021contrast}$_{\text{\demph{(IEEE TIP 22)}}}$ & LSTM & - & - & 67.6 & 73.8 & 56.2 & 57.0 & 82.1  \\
AimCLR~\cite{guo2022contrastive}$_{\text{\demph{(AAAI 22)}}}$ & GCN & 1.15G & 0.85M & 74.3 & 79.7 & 63.4 & 63.4 & 83.4  \\
RVTCLR+~\cite{zhu2023modeling}$_{\text{\demph{(ICCV 23)}}}$ & GCN & - & 0.85M & 74.7 & 79.1 & - & - & - \\
PSTL~\cite{zhou2023self}$_{\text{\demph{(AAAI 23)}}}$ & GCN & 1.15G & 0.85M & 77.3 & 81.8 & 66.2 & 67.7 & 88.4 \\
ViA~\cite{yang2024view}$_{\text{\demph{(IJCV 24)}}}$ & GCN & - & 0.85M & 78.1 & 85.8 & 69.2 & 66.9 & -  \\
HYSP~\cite{francohyperbolic}$_{\text{\demph{(ICLR 23)}}}$ & GCN & - & 0.85M & 78.2 & 82.6 & 61.8 & 64.6 & 83.8  \\
CPM~\cite{zhang2022contrastive}$_{\text{\demph{(ECCV 22)}}}$ & GCN & 2.22G & 0.85M & 78.7 & 84.9 & 68.7 & 69.6 & 88.8  \\
CMD~\cite{mao2022cmd}$_{\text{(ECCV 22)}}$ & BiGRU & 5.76G & 10.0M & 79.8 & 86.9 & 70.3 & 71.5 & -  \\ 
SkeAttnCLR~\cite{hua2023part}$_{\text{\demph{(IJCAI 23)}}}$ & GCN & 1.15G & 0.85M & 80.3 & 86.1 & 66.3 & 74.5 & 87.3  \\
HiCLR~\cite{zhang2023hierarchical}$_{\text{\demph{(AAAI 23)}}}$ & GCN & 1.15G & 0.85M & 80.4 & 85.5 & 70.0 & 70.4 & -  \\
HaLP~\cite{shah2023halp}$_{\text{\demph{(CVPR 23)}}}$ & BiGRU & - & 10.0M & 79.7 & 86.8 & 71.1 & 72.2 & -  \\ 
ActCLR~\cite{lin2023actionlet}$_{\text{\demph{(CVPR 23)}}}$ & GCN & 1.15G & 0.84M & 80.9 & 86.7 & 69.0 & 70.5 & -  \\ 
UmURL~\cite{sun2023unified}$_{\text{\demph{(ACM MM 23)}}}$ & Transformer & 1.74G & 20.3M & 82.3 & 89.8 & 73.5 & 74.3 & -  \\
Eq-Contrast~\cite{lin2024mutual}$_{\text{\demph{(IEEE TIP 24)}}}$ & BiGRU & 2.88G & 10.0M & 83.9 & 90.3 & 75.7 & 77.2 & 89.7 \\
PCM$^3$~\cite{zhang2023prompted}$_{\text{\demph{(ACM MM 23)}}}$ & BiGRU & 2.88G & 10.0M & 83.9 & 90.4 & 76.5 & 77.5 & -  \\
MAMP~\cite{mao2023masked}$_{\text{\demph{(ICCV 23)}}}$ & Transformer & - & - & 84.9 & 89.1 & \underline{78.6} & 79.1 & \underline{92.2}  \\
SCD-Net~\cite{wu2024scd}$_{\text{\demph{(AAAI 24)}}}$ & GCN\&Transformer & 7.14G & 83.26M & \underline{86.6} & \underline{91.7} & 76.9 & \underline{80.1} & 91.9  \\ \hline
\multirow{4}{*}{${\text A}^2$MC (Ours)}& Transformer & 4.12G & 3.12M & 77.4 & 84.7 & 69.9 & 71.7 & 86.2  \\
& GCN & 1.21G & 0.85M & 78.8 & 85.1 & 69.5 & 71.3 & 86.1  \\
& BiGRU & 2.93G & 10.0M & 80.0 & 87.9 & 71.4 & 73.2 & 87.4  \\
& GCN\&Transformer & 7.33G & 83.26M & \bf{86.7} & \bf{91.8} & \bf{79.6} & \bf{80.4} & \bf{92.5}  \\
\hline
\end{tabular}
\end{table*}

\noindent\textbf{Self-supervised Pre-training.}
In the contrastive learning based on MoCo v2~\cite{chen2020improved}, the size of memory bank $K$, the temperature hyper-parameter $\tau$, and momentum coefficient $\alpha$ are set to 16,384, 0.07, and 0.999, respectively. On NTU-60 and NTU-120, $\text{A}^2$MC is pre-trained for 450 total epochs, and the learning rate drops to 0.001 after 350 epochs. On PKU-\Rmnum{2}, $\text{A}^2$MC is pre-trained for 1,000 total epochs, and the learning rate drops to 0.001 after 800 epochs. In Att-Aug, the learning rate $\epsilon$ and the scalar value $\eta$ are set to 0.1 and 0.5 for the one-step gradient-based White-box Attack. For the learnable memory bank $M_0$ of PNM, the learning rate $\beta$ is 3.0 with a lower temperate of 0.03. To obtain more hard negative features, different mixing coefficients ($\lambda=0.4,0.3,0.2,0.1$) are set for pre-training.

\noindent\textbf{Linear Evaluation.} A learnable linear classifier is attached to the frozen query encoder to predict action classes. The network is trained for 80 epochs with an initial learning rate of 0.1 (drops to 0.01 and 0.001 at epoch 50 and 70).

\noindent\textbf{KNN Evaluation.} A K-Nearest Neighbor
(KNN) classifier is directly used to classify test samples without training, based on the K-nearest neighbor of features of the training samples learned by the query encoder, following~\cite{su2020predict}.


\noindent\textbf{Transfer Learning Evaluation.} The query encoder is first pre-trained on the source dataset, and then fine-tuned together with the linear classifier on the target dataset for 80 epochs with an initial learning rate of 0.01 (drops to 0.001 and 0.0001 at epoch 50 and 70). The source dataset is the training set from NTU-60/NTU-120 (x-sub), while the target dataset is the testing set from PKU-\Rmnum{2}. 

\renewcommand{\arraystretch}{1.3}
\begin{table}
\caption{Linear evaluation results in different modalities on NTU-60. J, M, and B refer to the three modalities: joint, motion, and bone, respectively.}
\label{modality}
\centering    
  \setlength{\tabcolsep}{8pt}
\begin{tabular}{l|c|c}
\hline
Method & Modality & x-sub \\ \hline\hline
${\text A}^2$MC & J & 80.0 \\
${\text A}^2$MC & M & 77.1 \\
${\text A}^2$MC & B & 78.3 \\ \hline
3s-CrosSCLR~\cite{li20213d} & J+M+B & 77.8 \\
3s-AimCLR~\cite{guo2022contrastive} & J+M+B & 78.9 \\
3s-RVTCLR+~\cite{zhu2023modeling} & J+M+B & 79.7 \\
3s-HiCLR~\cite{zhang2022hierarchical} & J+M+B & 80.4 \\
3s-CrosSCLR-B~\cite{li20213d} & J+M+B & 82.1 \\
3s-CPM~\cite{zhang2022contrastive} & J+M+B & 83.2 \\
3s-HiCo~\cite{dong2022hierarchical} & J+M+B & 83.8 \\
3s-CMD~\cite{mao2022cmd} & J+M+B & 84.1 \\
3s-ActCLR~\cite{lin2023actionlet} & J+M+B & 84.3 \\
3s-${\text A}^2$MC & J+M+B & \bf{84.6} \\
\hline
\end{tabular}
\end{table}

\subsection{Comparison with State-of-the-arts}
\noindent\textbf{Linear Evaluation Results.} 
As shown in~\ref{li_compare}, the proposed ${\text A}^2$MC achieves the best performance compared with other related works. Notably, ${\text A}^2$MC with encoder BiGRU performs best on most benchmarks (NTU-60 x-view, NTU-120 x-sub/x-set).
%
Specifically, with the consistent BiGRU configuration, ${\text A}^2$MC outperforms the corresponding SOTA method (i.e., CMD~\cite{mao2022cmd}) by 1.7\% and 5.4\% on NTU-120 (x-set) and PKU-\Rmnum{2} (x-sub). Here, CMD mines positive pairs among existing negative pairs based on the neighboring similarity distributions, while ${\text A}^2$MC constructs new hard positive features by attacking and augmenting skeletons. ${\text A}^2$MC outperforms method HaLP~\cite{shah2023halp} overall, both generate hard positive pairs, but ${\text A}^2$MC also generates hard negative pairs based on hard positive pairs.
With the same GCN (i.e., ST-GCN~\cite{yan2018spatial}) configuration, ${\text A}^2$MC is comparable to CPM~\cite{zhang2022contrastive} and ActCLR~\cite{lin2023actionlet}. 
The ${\text A}^2$MC considers the semantic-attacked itself as the hard positive feature, while CPM~\cite{zhang2022contrastive} identifies others in a contextual queue as the hard positive ones, and ActCLR~\cite{lin2023actionlet} treats skeletons resulting from distinct augmentations in both motion and static parts as more diverse samples.
 With Transformer (i.e., DSTA-Net~\cite{shi2020decoupled}) as encoder, ${\text A}^2$MC also achieves better performance with fewer parameters compared to Transformer-based method, i.e., GL-Transformer~\cite{kim2022global}.
Compared to the SOTA method SCD-Net~\cite{wu2024scd}, which utilizes both GCN \& Transformer, our approach achieves comparable performance with the same backbone, and even outperforms it on NTU-120 and PKU-I.

We also conduct comparative experiments to evaluate the performance of  ${\text A}^2$MC using the three-modality data (including joint, motion, and bone) as input on NTU-60 (x-sub). For multi-modal skeleton sequences with joint, motion, and bone, we simply extend ${\text A}^2$MC to a three-stream version, i.e., 3s-${\text A}^2$MC, without any elaborate design. As shown in~\ref{modality}, 3s-${\text A}^2$MC achieves comparable performance to that of 3s-CMD~\cite{mao2022cmd} and 3s-ActCLR~\cite{lin2023actionlet}, which is the SOTA method designed for specifically handling multiple modalities.

\noindent\textbf{KNN Evaluation Results.} KNN Evaluation is also regarded as the action retrieval task in some previous works~\cite{dong2022hierarchical,thoker2021skeleton}. From the KNN evaluation results in~\ref{knn}, ${\text A}^2$MC outperforms the SOTA method (i.e., CMD~\cite{mao2022cmd}) by 1.7\% on NTU-120, and it is comparable to CMD on NTU-60. It demonstrates that  ${\text A}^2$MC learns high-quality features in the discriminative space in an unsupervised way.

\renewcommand{\arraystretch}{1.3}
\begin{table}
\caption{KNN evaluation results on NTU-60, and NTU-120.}
\label{knn}
\centering    
  \setlength{\tabcolsep}{6pt}
\begin{tabular}{l|c|c|c|c}
\hline
\multirow{2}{*}{Method}&
\multicolumn{2}{c|}{NTU-60}&
\multicolumn{2}{c}{NTU-120}\\
\cline{2-5}
& x-sub & x-view & x-sub & x-set \\ \hline\hline
LongT GAN~\cite{zheng2018unsupervised} & 39.1 & 48.1 & 31.5 & 35.5 \\
P$\&$C~\cite{su2020predict} & 50.7 & 76.3 & 39.5 & 41.8 \\
ISC~\cite{thoker2021skeleton} & 62.5 & 82.6 & 50.6 & 52.3 \\
\dempg{CrosSCLR-B}~\cite{li20213d} & \dempg{66.1} & \dempg{81.3} & \dempg{52.5} & \dempg{54.9} \\
HiCLR~\cite{zhang2022hierarchical} & 67.3 & 75.3 & - & - \\
HaLP~\cite{shah2023halp} & 65.8 & 83.6 & 55.8 & 59.0 \\
HiCo~\cite{dong2022hierarchical} & 68.3 & 84.8 & 56.6 & 59.1 \\
\dempg{CMD}~\cite{mao2022cmd} & \dempg{70.6} & \dempg{85.4} & \dempg{58.3} & \dempg{60.9} \\ \hline
${\text A}^2$MC (Ours) & \bf{70.8} & \bf{85.4} & \bf{59.1} & \bf{62.6} \\
\hline
\end{tabular}
\end{table}

\noindent\textbf{Transfer Learning Evaluation Results.} We conduct experiments to test whether the knowledge learned from the source dataset is helpful for learning on the target dataset, i.e.,, whether the learned features are generalizable.  The t
ransfer learning evaluation results are shown in~\ref{transfer}. Compared with other methods, the features learned by ${\text A}^2$MC show better generalization, especially with 2.1\% higher performance than CMD~\cite{mao2022cmd} when transferring from NTU-60 (x-sub) to PKU-\Rmnum{2}.

\renewcommand{\arraystretch}{1.3}
\begin{table}
\caption{Transfer learning evaluation results.}
\label{transfer}
\begin{center}
\begin{tabular}{l|c|c}
\hline
\multirow{2}{*}{Method}&
\multicolumn{2}{c}{Transfer to PKU-\Rmnum{2}}\\
\cline{2-3}
& NTU-60 (x-sub) & NTU-120 (x-sub) \\ \hline\hline
LongT GAN~\cite{zheng2018unsupervised} & 44.8 & - \\
MS$^2$L~\cite{lin2020ms2l} & 45.8 & - \\
ISC~\cite{thoker2021skeleton} & 51.1 & 52.3 \\
\dempg{CrosSCLR-B}~\cite{li20213d} & \dempg{54.0} & \dempg{52.8} \\
\dempg{CMD}~\cite{mao2022cmd} & \dempg{56.0} & \dempg{57.0} \\ \hline
${\text A}^2$MC (Ours) & \bf{58.1} & \bf{58.9} \\
\hline
\end{tabular}
\end{center}
\end{table}

\subsection{Ablation Study}

\noindent\textbf{Effectiveness of different components.} To demonstrate the effectiveness of different components (including Att-Aug, PNM, and MC Loss) in the ${\text A}^2$MC framework, we conduct the ablation study for different components on NTU-60, as shown in~\ref{ablation}. Based on B1 (basic contrastive learning with basic augmentations and InfoNCE loss), B3 (adding Att-Aug) improves performance by 2.9\% and 2.8\% on x-sub and x-view, proving the superiority of Att-Aug. B2 (with PNM alone) improves by 1.3\% and 1.4\% compared to B1, confirming that~\cite{hu2021adco} is applicable to skeletons. By replacing InfoNCE loss with MC Loss in B4, the performance is further improved. The difference between InfoNCE loss and MC Loss is that the latter pulls the feature distribution toward the similarity distribution calculated from positive features instead of the one-hot distribution. Finally, by adding PNM based on B4, ${\text A}^2$MC achieves the best performance that shows the effectiveness of PNM.

\renewcommand{\arraystretch}{1.3}
\begin{table}
\caption{Ablation study under linear evaluation on NTU-60.}
\label{ablation}
\centering    
  \setlength{\tabcolsep}{5pt}
\begin{tabular}{c|c|c|c|c|c}
\hline
\multirow{2}{*}{Baseline}& \multirow{2}{*}{Att-Aug}&
\multirow{2}{*}{PNM}&
\multirow{2}{*}{MC}&
\multicolumn{2}{c}{NTU-60}\\
\cline{5-6}
& & & & x-sub & x-view \\ \hline\hline
 B1& {\color{lightgray}\XSolidBrush} & {\color{lightgray}\XSolidBrush} & {\color{lightgray}\XSolidBrush} & 75.6 & 82.8 \\
 B2& {\color{lightgray}\XSolidBrush} & \Checkmark & {\color{lightgray}\XSolidBrush} & 76.9 & 84.2 \\
 B3&\Checkmark & {\color{lightgray}\XSolidBrush} & {\color{lightgray}\XSolidBrush} & 78.5 & 85.6 \\
 B4&\Checkmark & {\color{lightgray}\XSolidBrush} & \Checkmark & 79.0 & 86.7 \\
${\text A}^2$MC&\Checkmark & \Checkmark & \Checkmark & \bf{80.0} & \bf{87.9} \\
\hline
\end{tabular}
\end{table}

\noindent\textbf{Effectiveness of targeted and untargeted augment.} We test the effects of the targeted semantic attack (Att) and the untargeted appearance augmentation (Aug) within the Att-Aug module. \ref{ablation_AttAug} demonstrates significant improvements when Att or Aug is added separately, with Att having a greater impact. This is attributed to the generation of more challenging samples closer to the semantic boundary. Combining Att and Aug on NTU-60 (x-view) boosts performance by 1.2\% and 0.8\% compared to using Att and Aug alone, respectively. Thus, they are integrated into the final model as the Att-Aug module.

\begin{table}
\caption{Ablation study under linear evaluation of targeted attack and untargeted augmentation on NTU-60.}
\label{ablation_AttAug}
\centering    
  \setlength{\tabcolsep}{8pt}
\begin{tabular}{c|c|c|c|c}
\hline
\multirow{2}{*}{Baseline}& \multirow{2}{*}{Att}&
\multirow{2}{*}{Aug}&
\multicolumn{2}{c}{NTU-60}\\
\cline{4-5}
& & & x-sub & x-view \\ \hline\hline
 A1& {\color{lightgray}\XSolidBrush} & {\color{lightgray}\XSolidBrush} & 75.6 & 82.8 \\
 A2&{\color{lightgray}\XSolidBrush} & \Checkmark & 79.0 & 86.7 \\
 A3&\Checkmark & {\color{lightgray}\XSolidBrush} & 79.9 & 87.1 \\
${\text A}^2$MC&\Checkmark & \Checkmark & \bf{80.0} & \bf{87.9} \\
\hline
\end{tabular}
\end{table}

\begin{figure*}
\begin{center}
\includegraphics[width=0.98\linewidth]{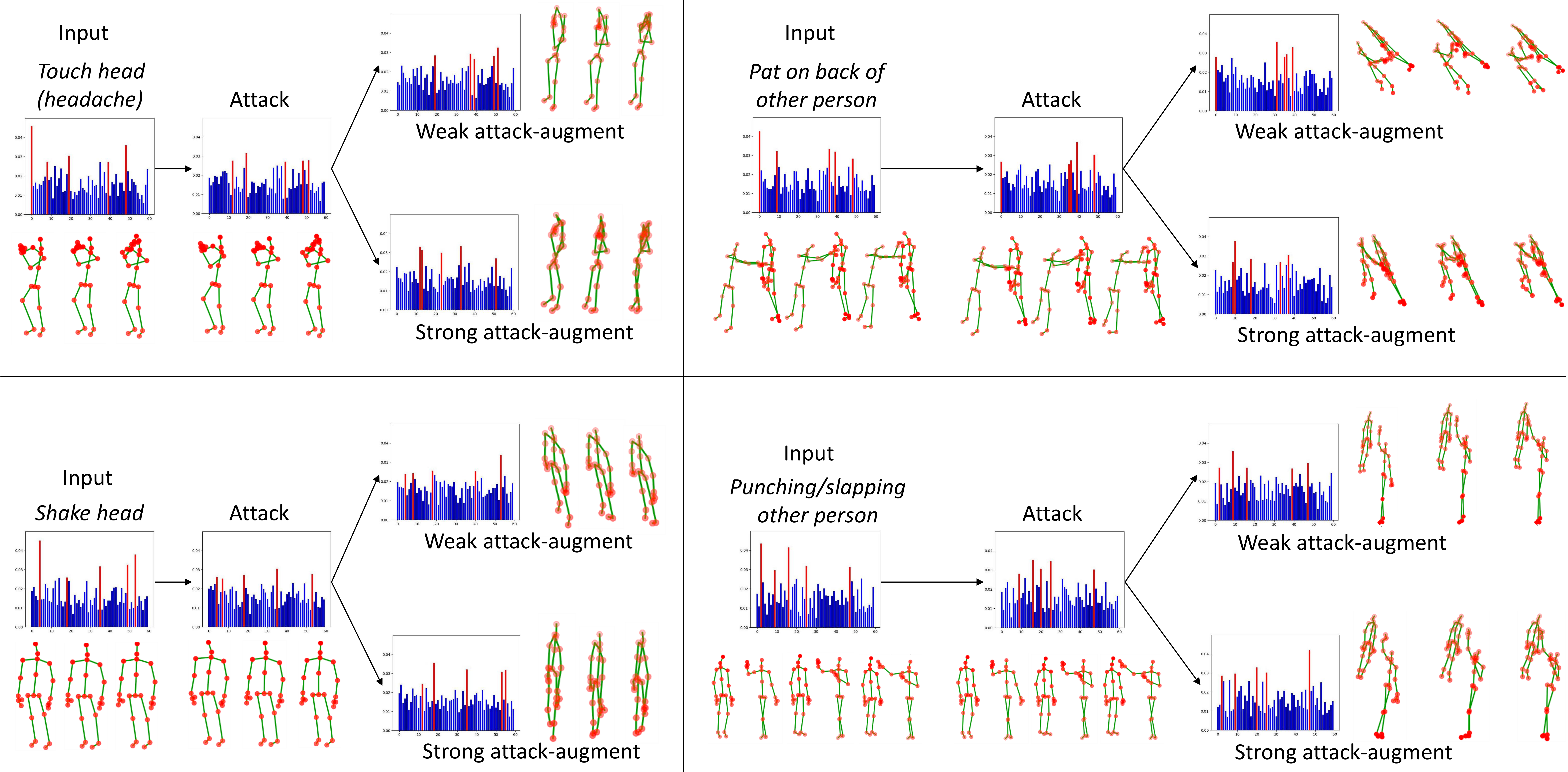}
\end{center}
   \caption{Visualization of attack-augmentation on NTU-60. For each group, given the original skeleton sequence, its skeletons and feature distributions after attack, weak attack-augmentation, and strong attack-augmentation are respectively visualized. Here, we utilize FC layer to obtain the feature distribution, where Top-5 bars are highlighted.}
\label{fig:attack}
\end{figure*}

\noindent\textbf{Visualization of Attack-Augmentation.} To intuitively illustrate the different effects of Attack-Augmentation (Att-Aug) at the semantic and appearance levels, we visualize the feature distributions and skeletons before and after the attack, weak attack-augmentation, and strong attack-augmentation, as shown in ~\ref{fig:attack}. From the perspective of the semantic level, the feature distribution after attack becomes smoother (refer to Top-5 bars) than before, 
while the results caused by weak/strong attack-augmentation have slight uncertainty.
 From the perspective of the appearance level, the changes of skeletons after attack are negligible, and the changes after strong attack-augmentation are relatively more obvious than those after weak attack-augmentation, which can further enrich representations with diverse appearance changes.


\begin{figure}[!t]
  \centering
  \subfloat[]{\includegraphics[width=0.45\linewidth]{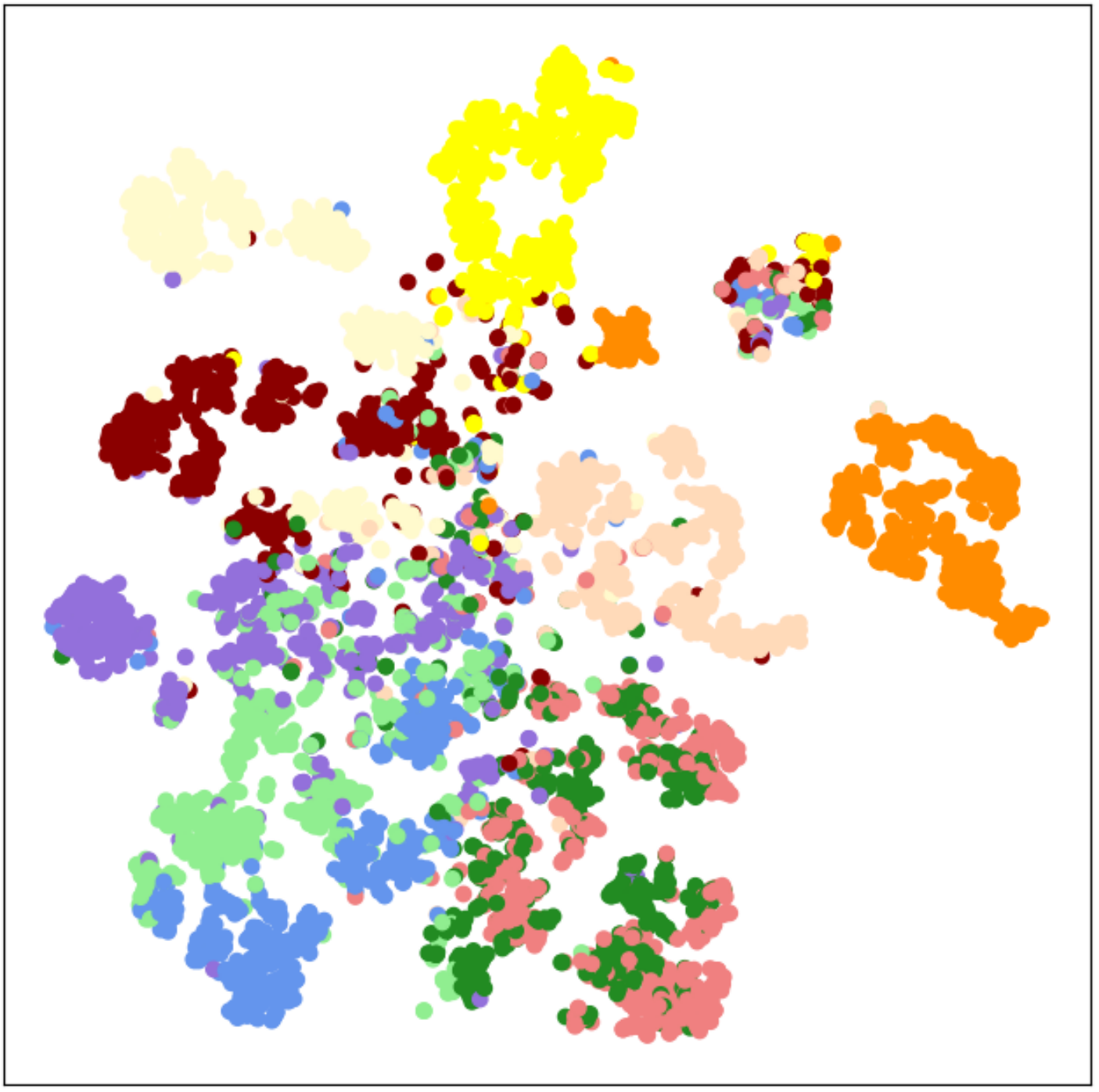}
  }
  \hfil
  \subfloat[]{\includegraphics[width=0.45\linewidth]{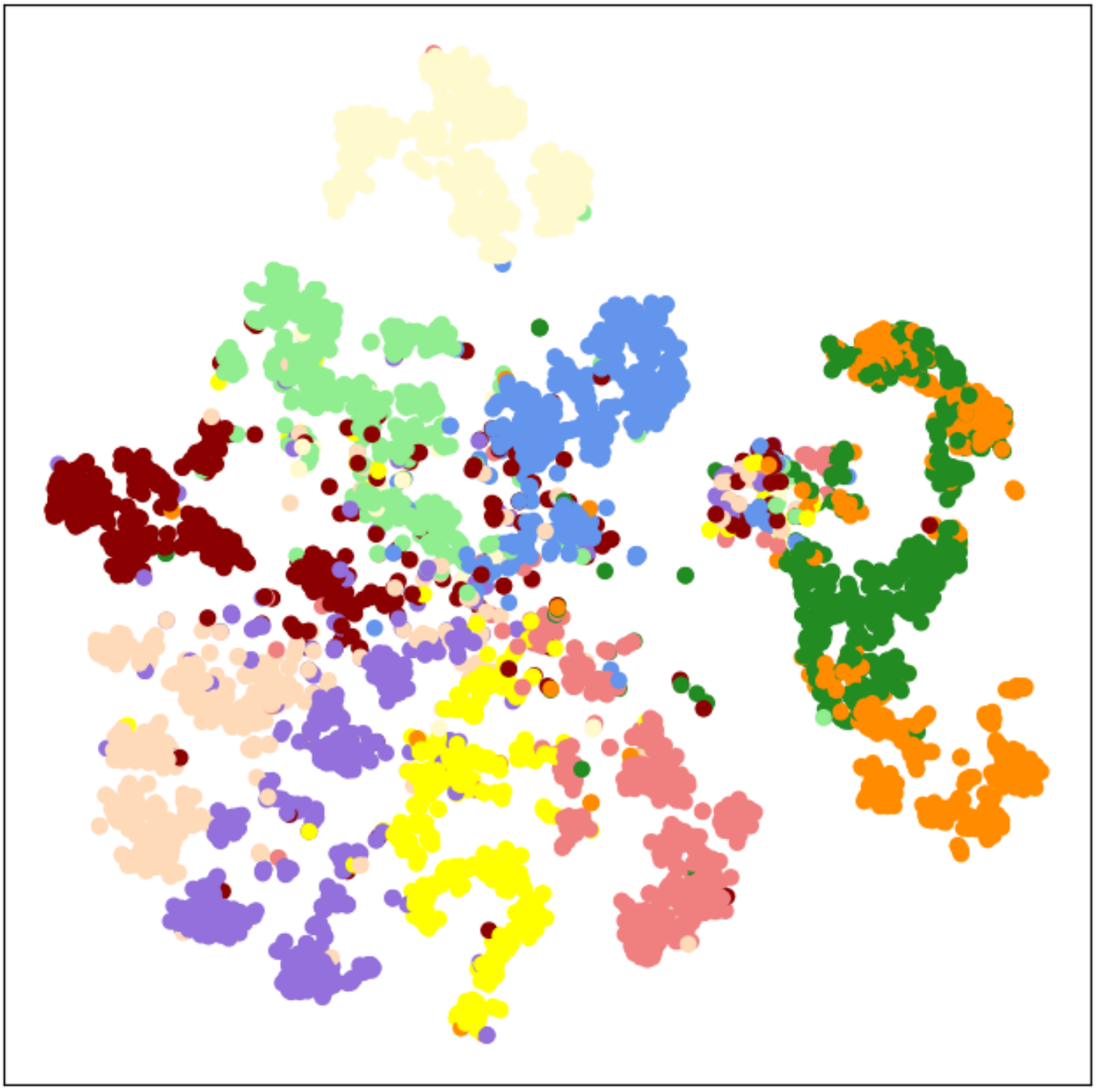}
  }
  \hfil
  \subfloat[]{\includegraphics[width=0.45\linewidth]{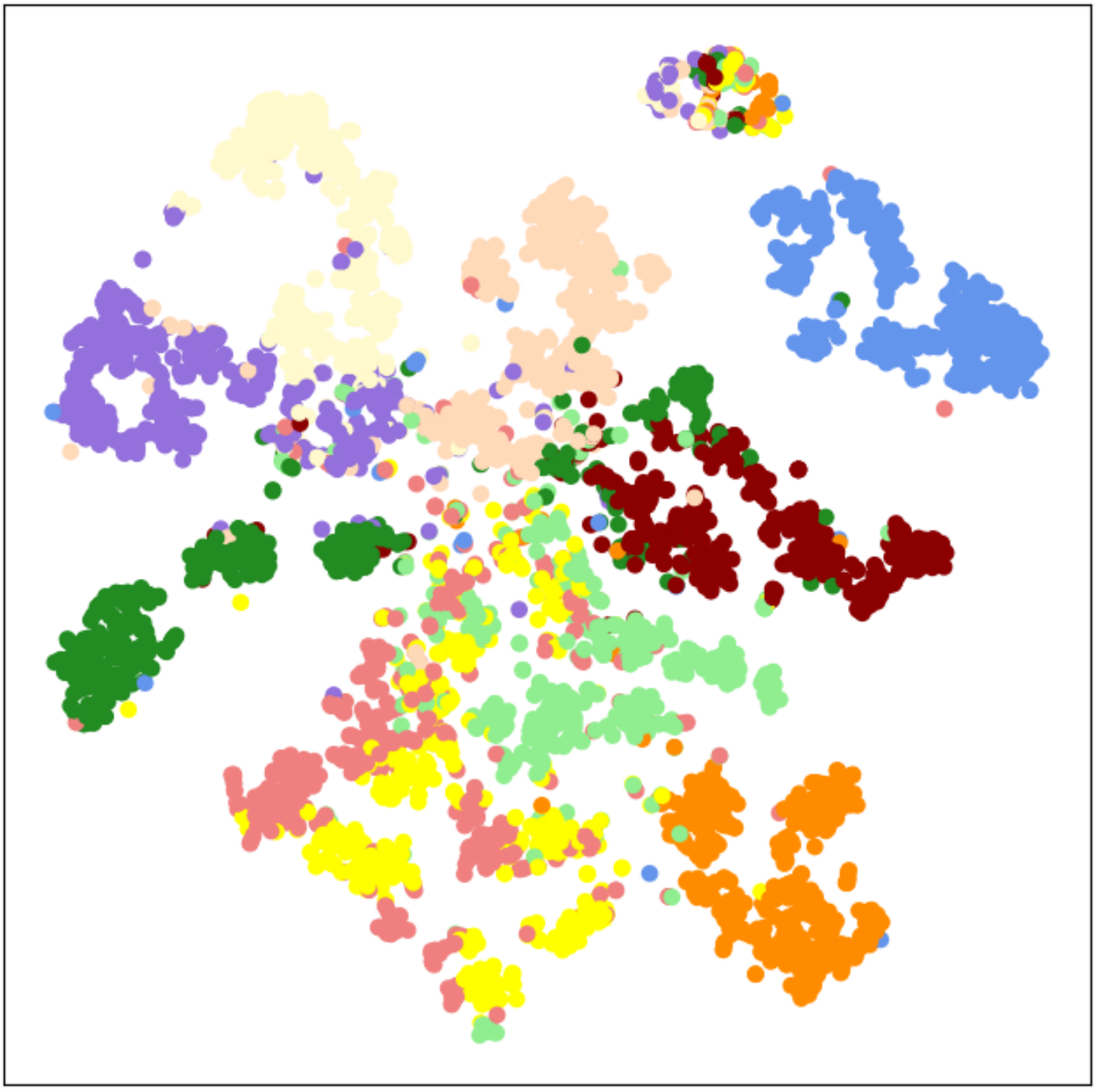}
  }
  \hfil
  \subfloat[]{\includegraphics[width=0.45\linewidth]{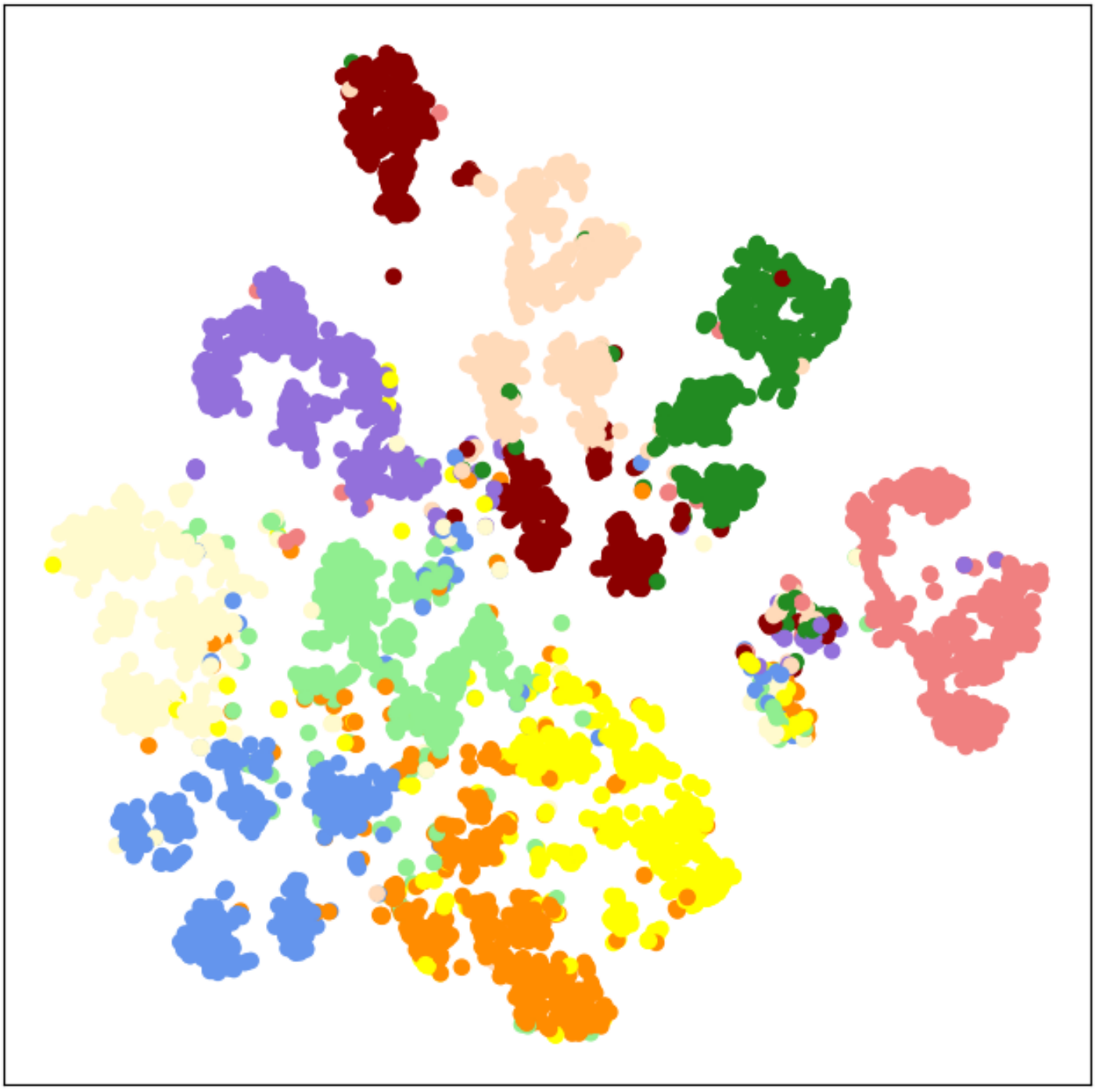}
  }
  \caption{The t-SNE on NTU-60. Ten action classes are randomly selected and
reported. (a) w/ weak. (b) w/ weak \& strong. (c) w/ attack. (d) ${\text A}^2$MC.}
  \label{fig_tsne}
\end{figure}

\noindent\textbf{The t-SNE Visualization of features.} We employ t-SNE algorithm to visualize the learned skeleton representations of ${\text A}^2$MC compared with those of three baselines, including ${\text A}^2$MC with only weak augmentation, ${\text A}^2$MC with only weak and strong augmentations, ${\text A}^2$MC with only attack. As shown in ~\ref{fig_tsne}, compared with such three baselines, the features learned by ${\text A}^2$MC are more aggregated in the same class and more discriminative in different classes, proving its representation learning capability.

\begin{figure}[!t]
  \centering
  \subfloat[]{\includegraphics[width=0.48\linewidth]{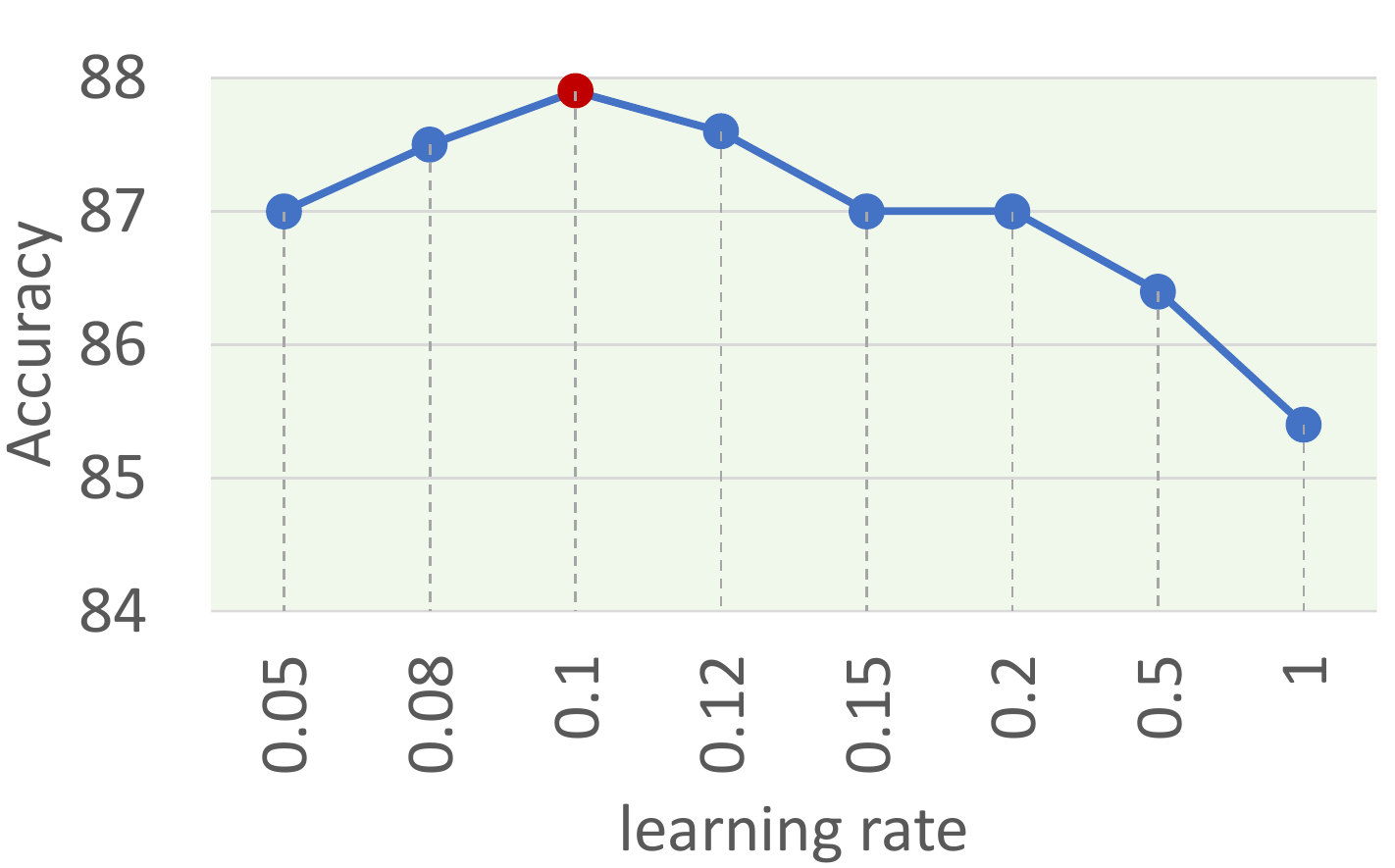}
  \label{fig:lr}}
  \hfil
  \subfloat[]{\includegraphics[width=0.48\linewidth]{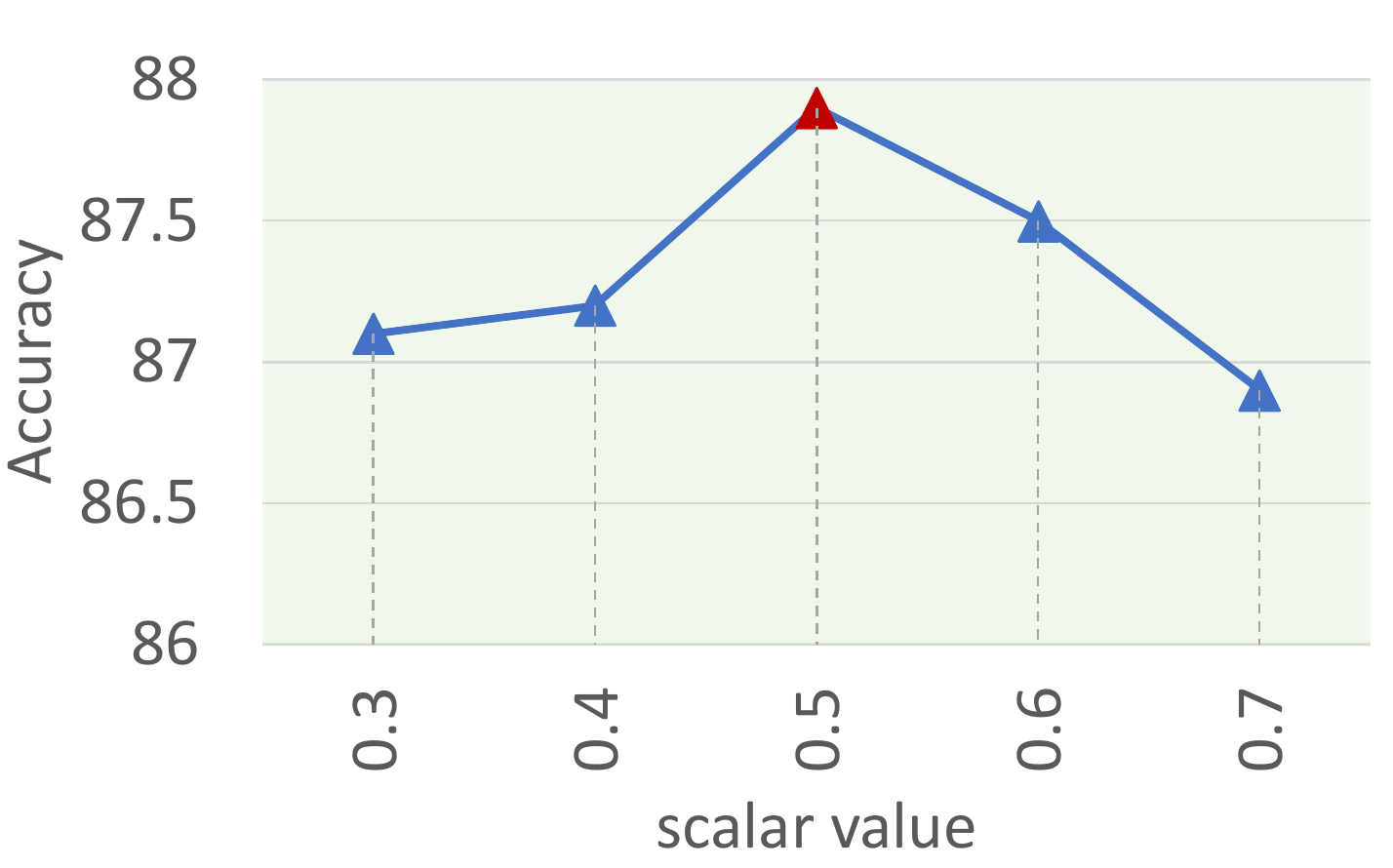}
  \label{fig:eta}}
  \caption{The linear evaluation results of different learning rate $\epsilon$ and scalar value $\eta$ in Att-Aug on NTU-60 (x-view). (a) The learning rate $\epsilon$. (b) The scalar value $\eta$.}
  \label{fig:lr_eta}
\end{figure}

\noindent\textbf{Effects of learning rate $\epsilon$ and scalar value $\eta$ in Att-Aug.}
The learning rate $\epsilon$ is an important parameter that affects perturbation in Att-Aug.
To investigate the effect of $\epsilon$, we conduct experiments with different values of $\epsilon$. As illustrated in Figure~\ref{fig:lr}, the best performance result on NTU-60 (x-view) is obtained when $\epsilon=0.1$. Therefore, $\epsilon$ is set to 0.1 in the final implementation. The scalar value $\eta$ as another key parameter that limits the perturbation amplitude produced by attack. Figure~\ref{fig:eta} shows that larger or smaller values can hurt performance, where large values may break the semantics and small values may weaken hard positive features. In all experiments, we choose $\eta=0.5$ as the default setting.

\begin{figure}[!t]
  \centering
  \subfloat[]{\includegraphics[width=0.48\linewidth]{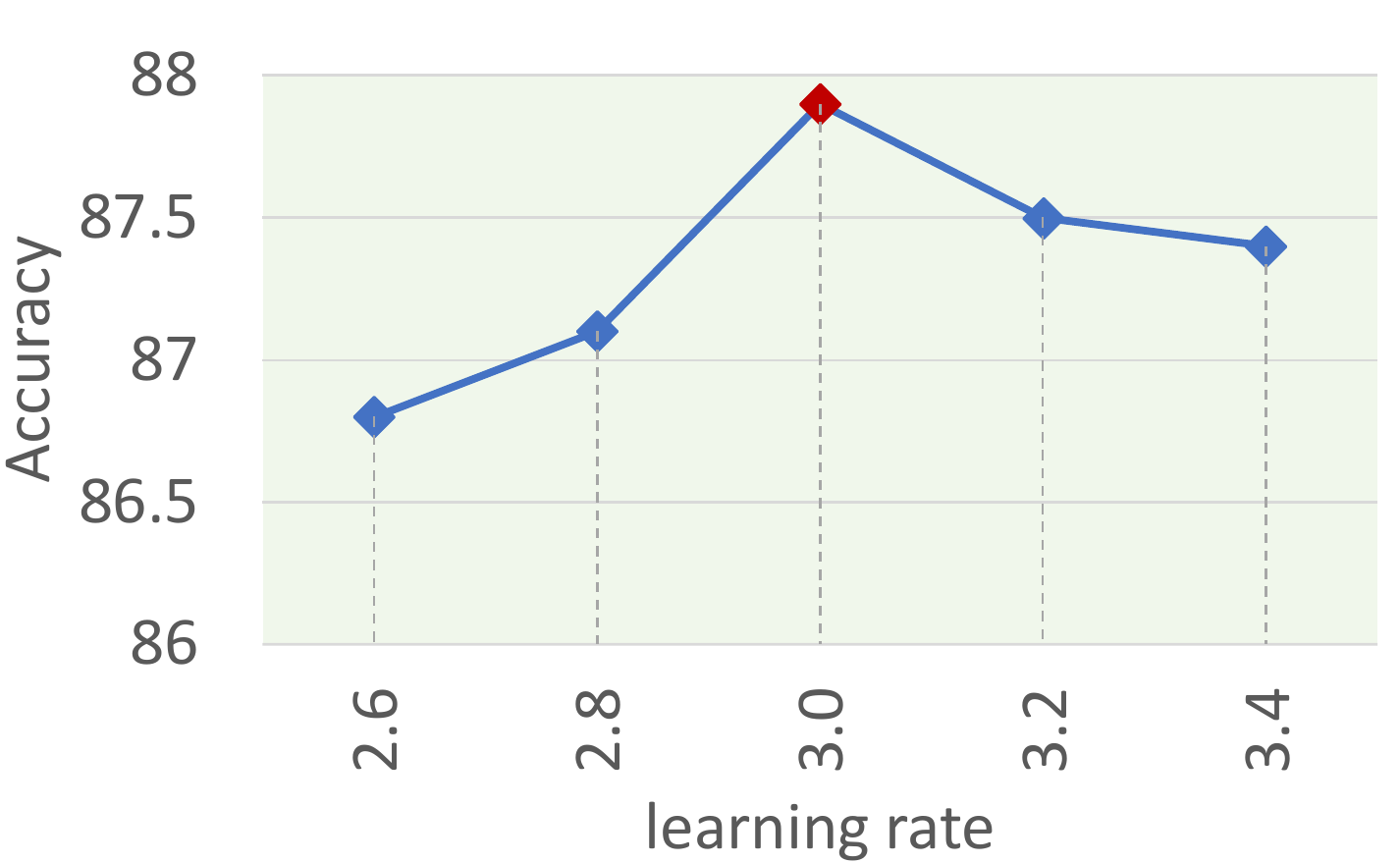}
  \label{fig:lr_neg}}
  \hfil
  \subfloat[]{\includegraphics[width=0.48\linewidth]{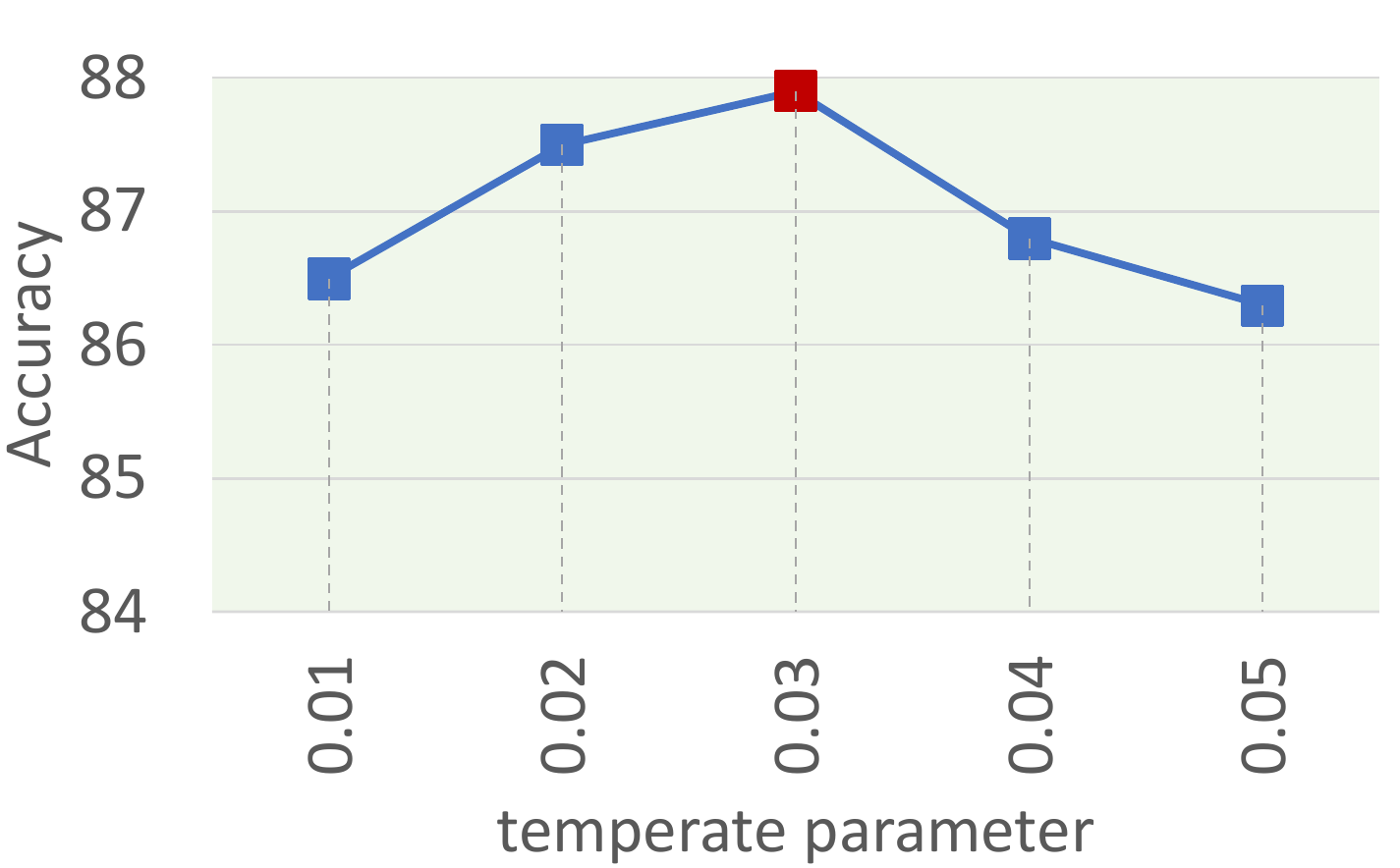}
  \label{fig:tau_neg}}
  \caption{The linear evaluation results of different learning rate $\beta$ and temperature in PNM on NTU-60 (x-view). (a) The learning rate $\beta$. (b) The temperature.}
  \label{fig:lr_tau_neg}
\end{figure}

\noindent\textbf{Effects of learning rate $\beta$ and temperature in PNM.}
In PNM, the learning rate $\beta$ with temperature is used to control the updating degree of the mixed memory banks~\cite{hu2021adco}. As shown in Figure~\ref{fig:lr_tau_neg}, it achieves the best downstream performance when $\beta=3.0$ with a temperature of 0.03.

\begin{figure}
\begin{center}
\includegraphics[width=0.85\linewidth]{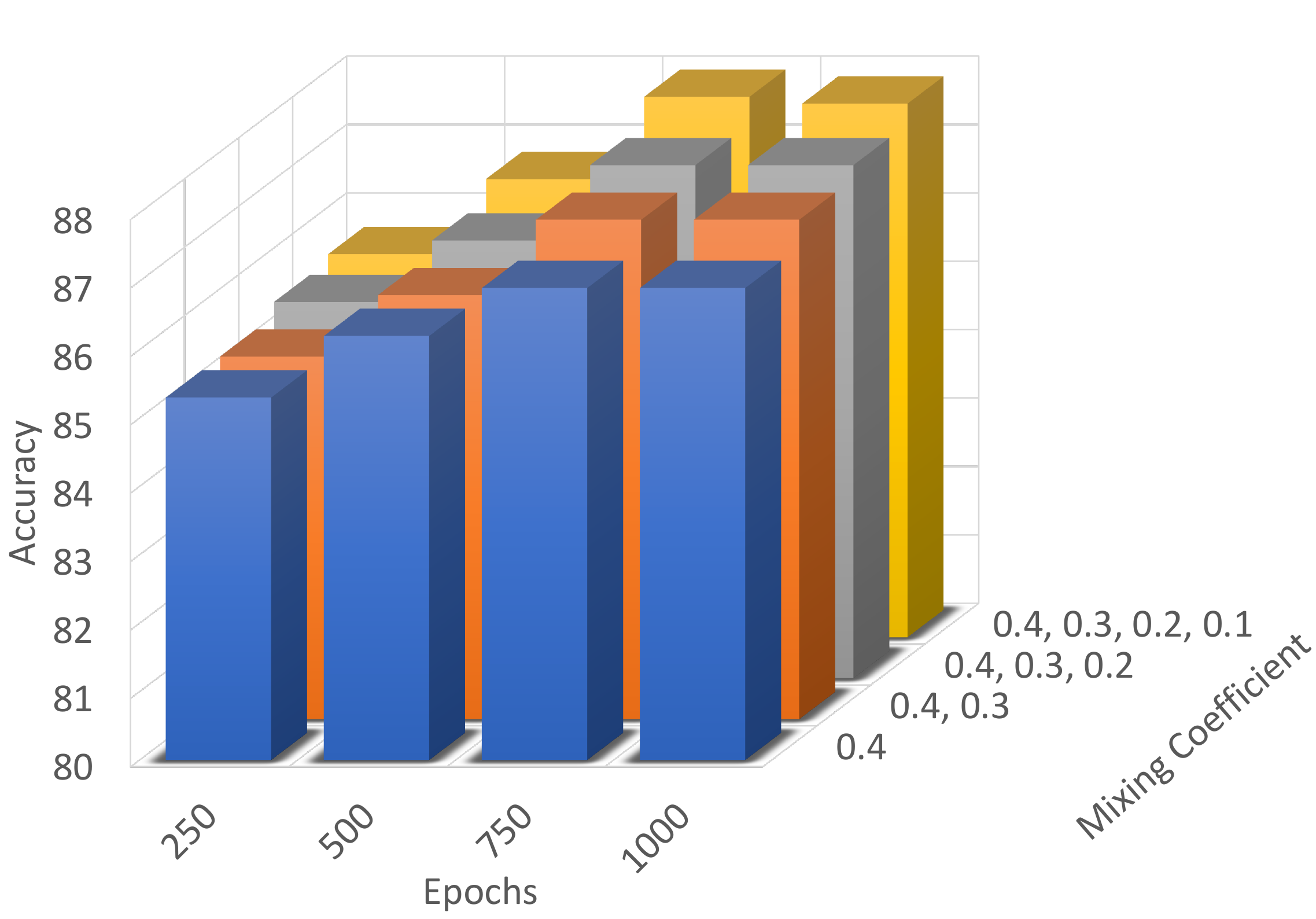}
\end{center}
   \caption{The linear evaluation results of different mixing coefficients $\lambda$ and epochs on NTU-60 (x-view).}
\label{fig:mix_epochs}
\end{figure}

\noindent\textbf{Effect of mixing coefficient $\lambda$ in PNM.}
To explore the effect of mixing coefficient $\lambda$, we test the performances by empirically setting four combined mixing coefficients, i.e., $\lambda\in\{(0.4), (0.4, 0.3), (0.4, 0.3, 0.2), (0.4, 0.3, 0.2, 0.1)\}$. Here, $\lambda=(0.4)$ means that this weight is adopted to mix positive-negative features for generating $K$ hard negative features, and $\lambda=(0.4, 0.3, 0.2, 0.1)$ means that these four weights are adopted to mix positive-negative features for generating 4$K$ hard negative features. As shown in Figure~\ref{fig:mix_epochs}, more hard negative features mixed by $\lambda=(0.4, 0.3, 0.2, 0.1)$ can help A$^2$MC to consistently yield better performance in different epochs. Therefore, $\lambda=(0.4, 0.3, 0.2, 0.1)$ is adopted in the final implementation.

\noindent\textbf{Performance with different epochs.}
To show the convergence of A$^2$MC with different numbers of hard negative features, the performance with different total epochs in the pre-training phase are reported.  As shown in Figure~\ref{fig:mix_epochs}, we observe that the performance is gradually improved from epoch 250 to epoch 750, finally stabilized at the epoch 750. More hard negative features may require more iterations compared with 450/300 total epochs of CMD~\cite{mao2022cmd}/CrosSCLR~\cite{li20213d} in related works.

\begin{table}
\caption{Finetuned evaluation results on NTU-60.}
\label{finetuned}
\centering    
  \setlength{\tabcolsep}{15pt}
\begin{tabular}{l|c|c}
\hline
\multirow{2}{*}{Method}&
\multicolumn{2}{c}{NTU-60}\\
\cline{2-3}
& x-sub & x-view \\ \hline\hline
ST-GCN~\cite{yan2018spatial} & 81.5 & 88.3 \\
SkeletonCLR~\cite{li20213d} & 82.2 & 88.9 \\
AimCLR~\cite{guo2022contrastive} & 83.0 & 89.2 \\
SkeleMixCLR~\cite{chen2022contrastive} & 84.5 & 91.1 \\
CPM~\cite{zhang2022contrastive} & 84.8 & 91.1 \\ \hline
${\text A}^2$MC (Ours) & \bf{84.8} & \bf{92.3}\\
\hline
\end{tabular}
\end{table}

\begin{table}[t]
\caption{Semi-supervised evaluation results on NTU-60.}
\label{semi-supervised0}
\centering    
  \setlength{\tabcolsep}{10pt}
\begin{tabular}{l|c|c|c|c}
\hline
\multirow{2}{*}{Method}&
\multicolumn{2}{c|}{x-sub}&
\multicolumn{2}{c}{x-view}\\
\cline{2-5}
& 1\% & 10\% & 1\% & 10\% \\ \hline\hline
LongT GAN~\cite{zheng2018unsupervised} & 35.2 & 62.0 & - & - \\
MS$^2$L~\cite{lin2020ms2l} & 33.1 & 65.2 & - & - \\
ASSL~\cite{si2020adversarial} & - & 64.3 & - & 69.8 \\
ISC~\cite{thoker2021skeleton} & 35.7 & 65.9 & 38.1 & 72.5 \\
CrosSCLR-B~\cite{li20213d} & 48.6 & 72.4 & 49.8 & 77.0 \\
CMD~\cite{mao2022cmd} & 50.6 & 75.4 & 53.0 & 80.2 \\ \hline
${\text A}^2$MC (Ours) & \bf{52.0} & \bf{76.4} & \bf{53.5} & \bf{81.5} \\
\hline
\end{tabular}
\end{table}

\begin{table}[t]
\caption{More semi-supervised evaluation results on NTU-60.}
\label{semi-supervised}
\centering    
  \setlength{\tabcolsep}{10pt}
\begin{tabular}{l|c|c|c|c}
\hline
\multirow{2}{*}{Method}&
\multicolumn{2}{c|}{x-sub}&
\multicolumn{2}{c}{x-view}\\
\cline{2-5}
& 5\% & 20\% & 5\% & 20\% \\ \hline\hline
ASSL~\cite{si2020adversarial} & 57.3 & 68.0 & 63.6 & 74.7 \\
ISC~\cite{thoker2021skeleton} & 59.6 & 70.8 & 65.7 & 78.2 \\
CrosSCLR-B~\cite{li20213d} & 67.7 & 76.1 &  70.6 & 81.9 \\
CMD~\cite{mao2022cmd} & 71.0 & 78.7 & 75.3 & 84.3 \\ \hline
${\text A}^2$MC (Ours) & \bf{72.2} & \bf{79.4} & \bf{75.9} & \bf{85.4} \\
\hline
\end{tabular}
\end{table}

\begin{table}[t]
\caption{Linear evaluation results in three modalities (joint, motion, and bone) on NTU-60 and NTU-120.}
\label{thrmodality}
\centering    
  \setlength{\tabcolsep}{3pt}
\begin{tabular}{l|c|c|c}
\hline
Method & NTU-60 xview & NTU-120 xsub & NTU-120 xset \\ \hline\hline
3s-CrosSCLR~\cite{li20213d} & 83.4 & 67.9 & 66.7 \\
3s-AimCLR~\cite{guo2022contrastive} & 83.8 & 68.2 & 68.8 \\
3s-RVTCLR+~\cite{zhu2023modeling} & 84.6 & 60.0 & 68.9 \\
3s-HiCLR~\cite{zhang2022hierarchical} & 85.5 & - & - \\
3s-CrosSCLR-B~\cite{li20213d} & 89.2 & 71.6 & 73.4 \\
3s-CPM~\cite{zhang2022contrastive}  & 87.0 & 73.0 & 74.0 \\
3s-HiCo~\cite{dong2022hierarchical} & 90.8 & \bf{75.9} & \bf{77.3} \\
3s-CMD~\cite{mao2022cmd} & 90.9 & 74.7 & 76.1 \\
3s-ActCLR~\cite{lin2023actionlet} & 88.8 & 74.3 & 75.7 \\
3s-${\text A}^2$MC & \bf{91.2} & 75.5 & 77.1 \\
\hline
\end{tabular}
\end{table}

\begin{figure}[!t]
  \centering
  \subfloat[]{\includegraphics[width=0.48\linewidth]{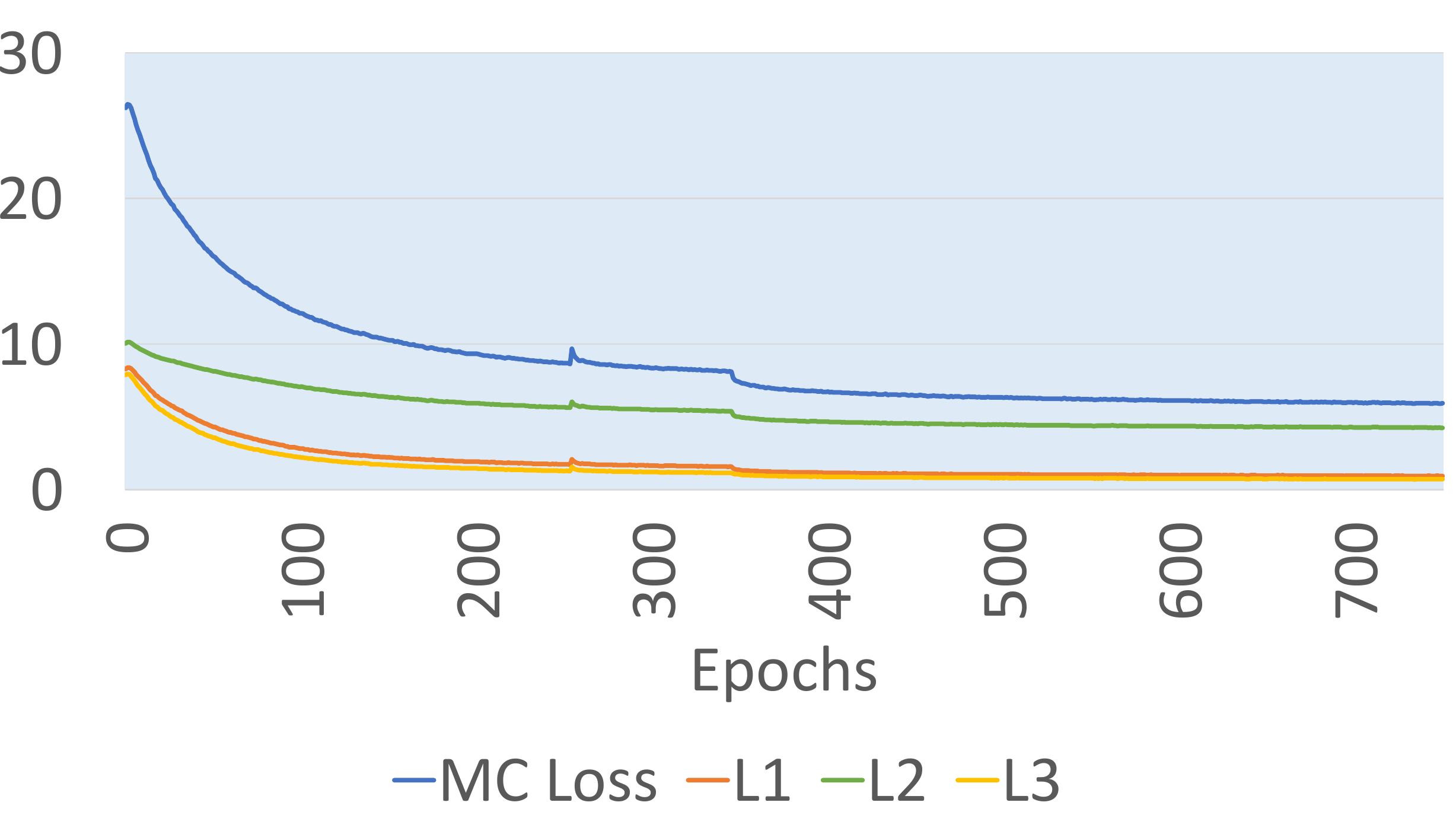}
  \label{NTUCSpre}}
  \hfil
  \subfloat[]{\includegraphics[width=0.48\linewidth]{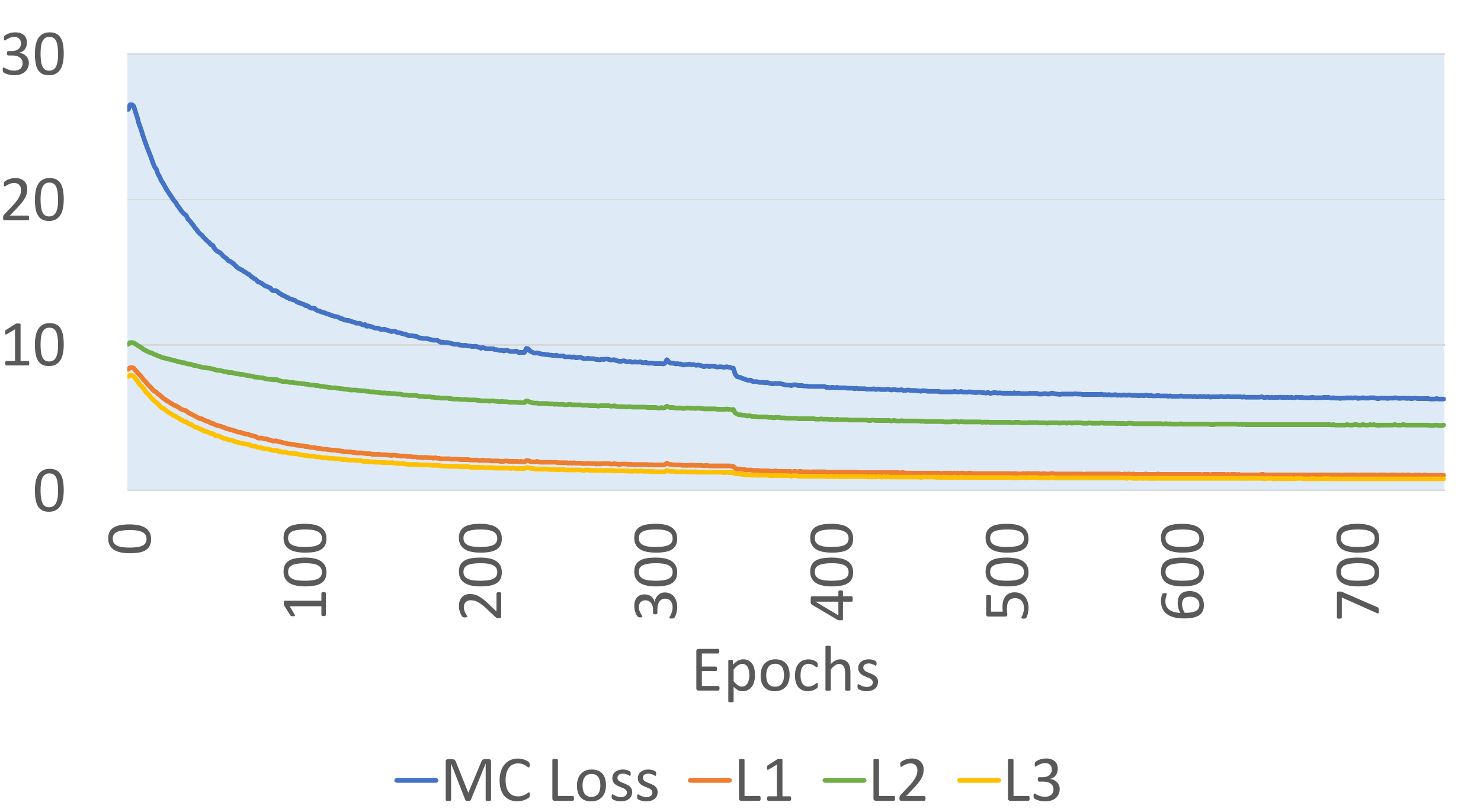}
  \label{NTUCVpre}}
  \hfil
  \subfloat[]{\includegraphics[width=0.48\linewidth]{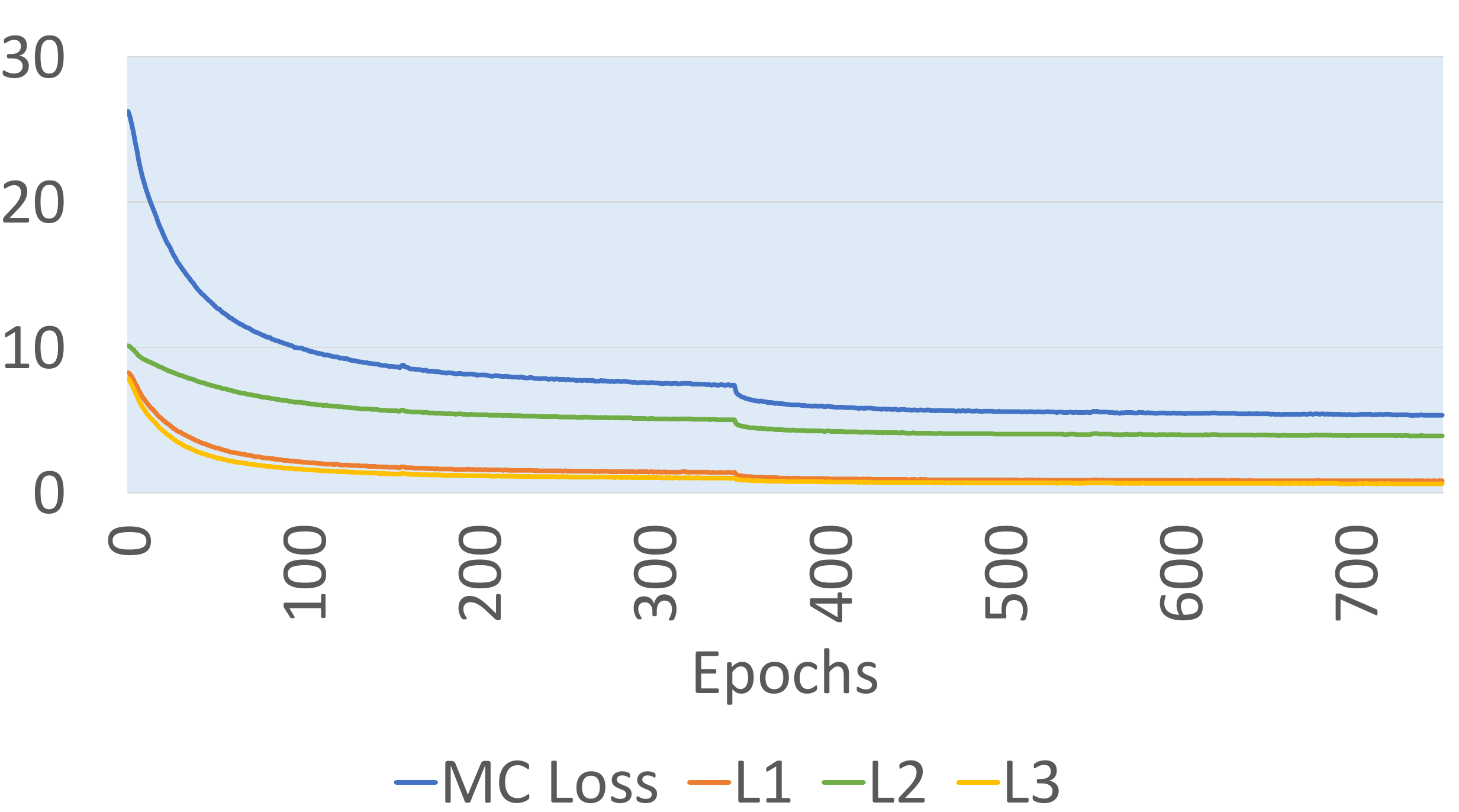}
  \label{NTU120CSpre}}
  \hfil
  \subfloat[]{\includegraphics[width=0.48\linewidth]{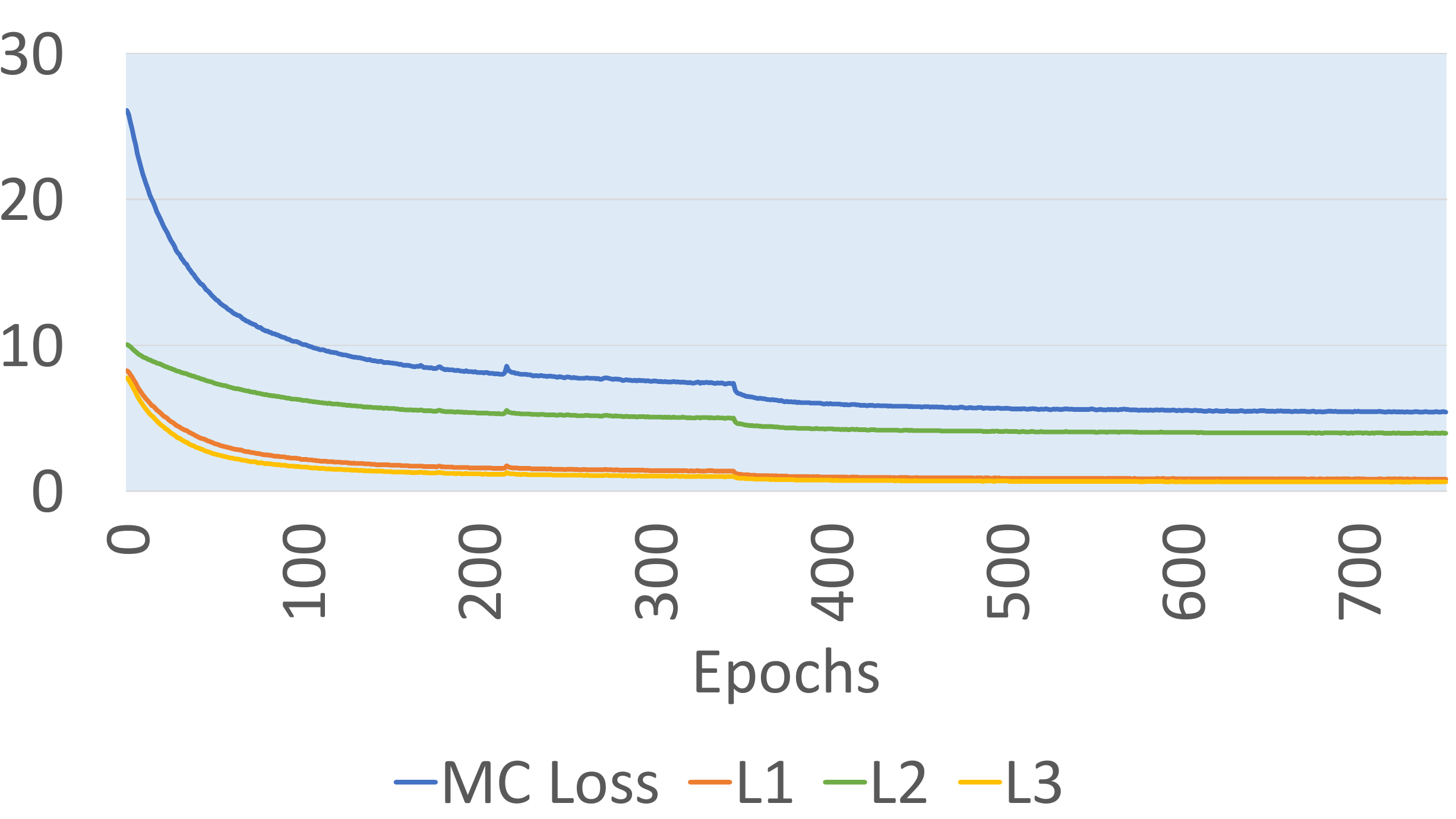}
  \label{NTU120CEpre}}
  \caption{The change curves of MC Loss, $\mathcal{L}_1$ loss, $\mathcal{L}_2$ loss, and $\mathcal{L}_3$ loss on NTU-60 and NTU-120 in the pre-training phase. (a) On NTU-60 (x-sub). (b) On NTU-60 (x-view). (c) On NTU-120 (x-sub). (d) On NTU-120 (x-set).}
  \label{fig_pre_curve}
\end{figure}

\begin{figure}[!t]
  \centering
  \subfloat[]{\includegraphics[width=0.48\linewidth]{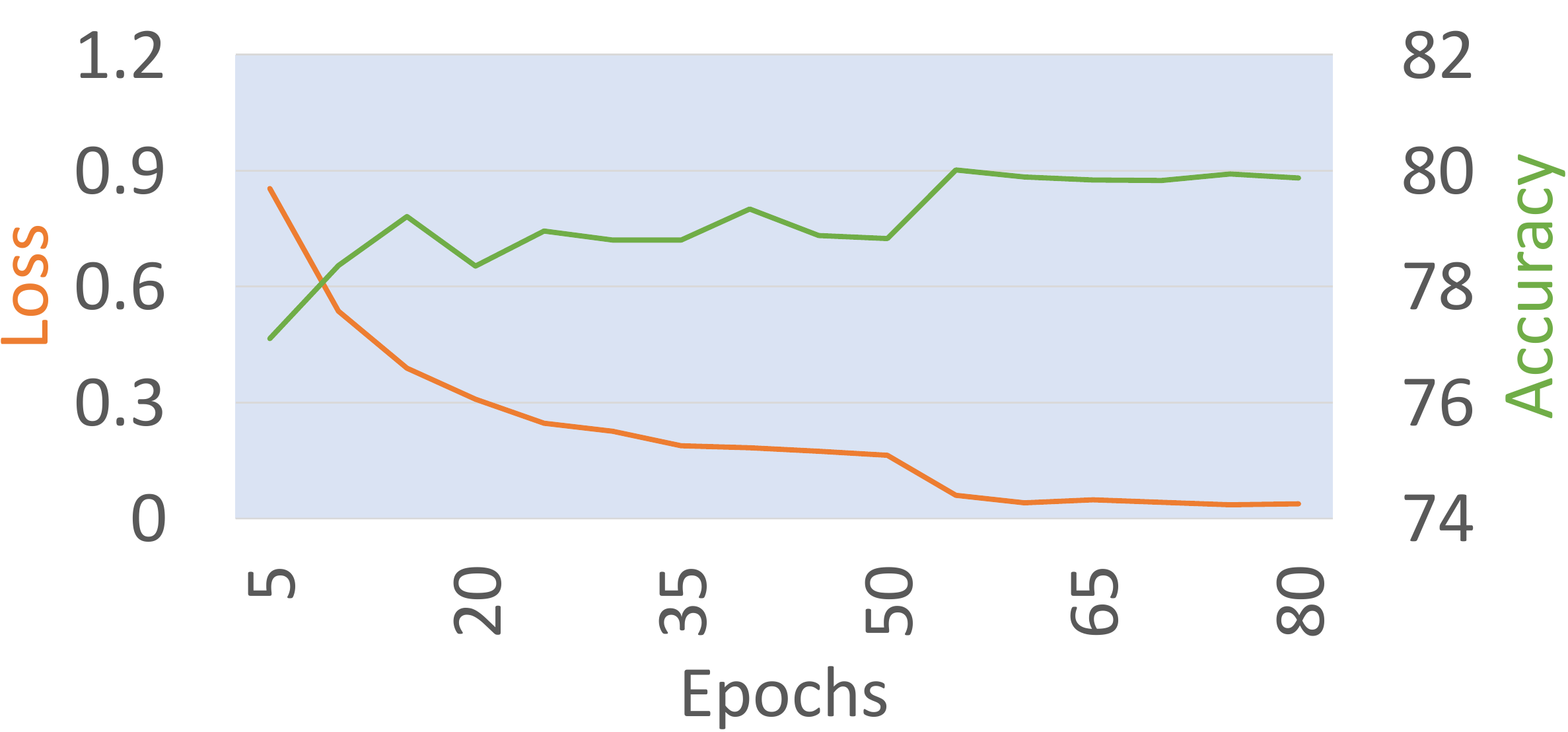}
  \label{NTUCSlin}}
  \hfil
  \subfloat[]{\includegraphics[width=0.48\linewidth]{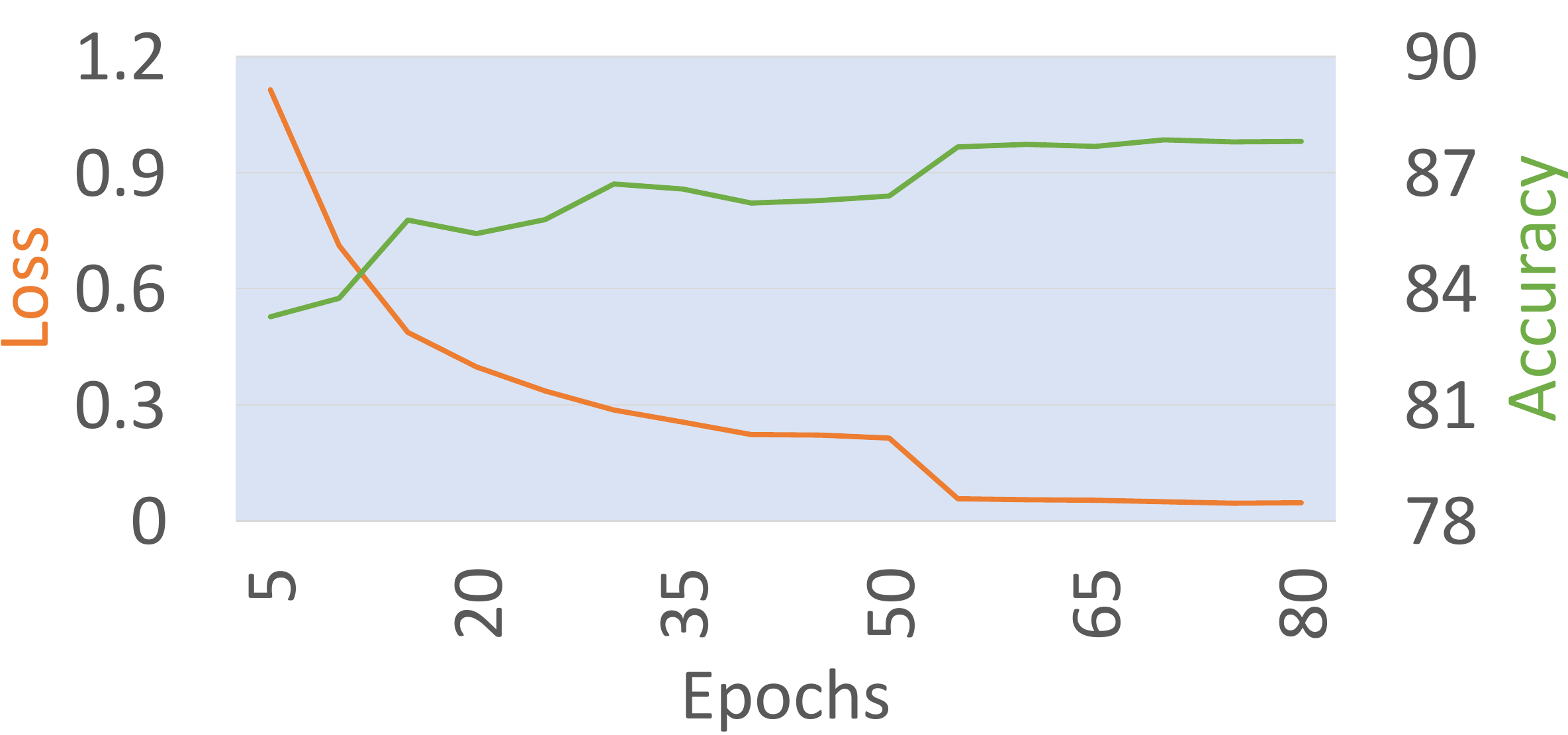}
  \label{NTUCVlin}}
  \hfil
  \subfloat[]{\includegraphics[width=0.48\linewidth]{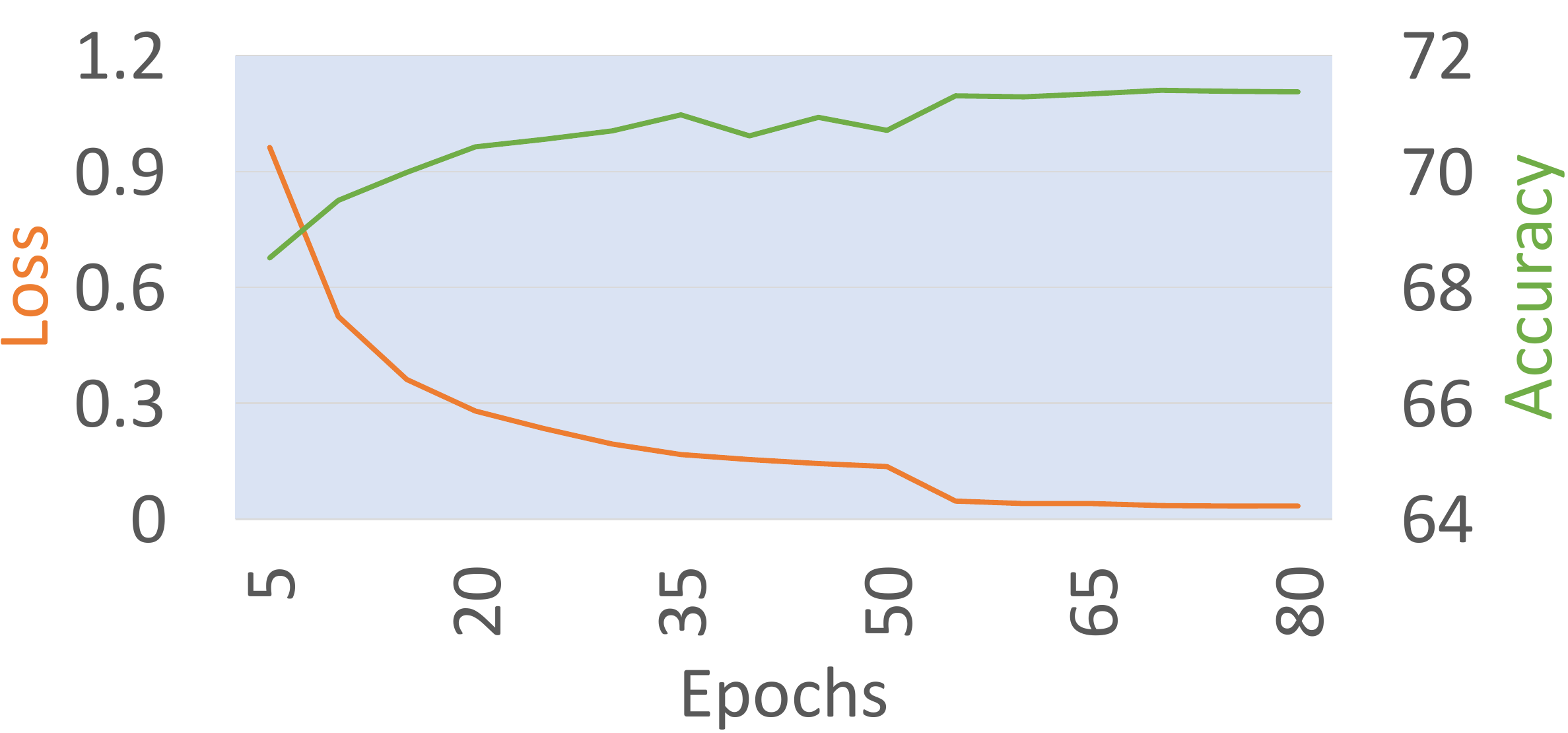}
  \label{NTU120CSlin}}
  \hfil
  \subfloat[]{\includegraphics[width=0.48\linewidth]{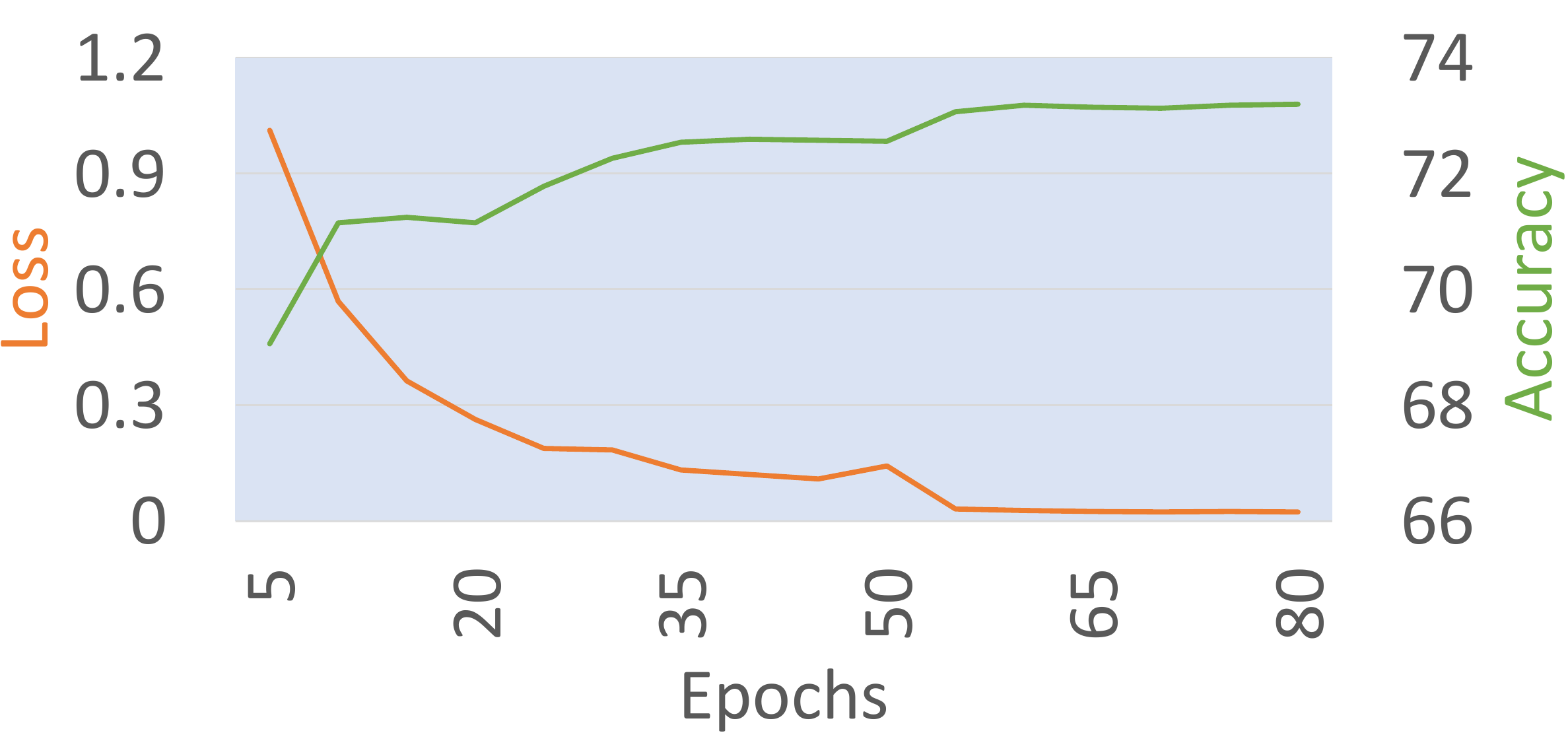}
  \label{NTU120CElin}}
  \caption{The change curves of recognition loss and accuracy on NTU-60 and NTU-120 in the linear evaluation. (a) On NTU-60 (x-sub). (b) On NTU-60 (x-view). (c) On NTU-120 (x-sub). (d) On NTU-120 (x-set).}
  \label{fig_lin_curve}
\end{figure}

\begin{figure}[!t]
  \centering
  \subfloat[]{\includegraphics[width=0.45\linewidth]{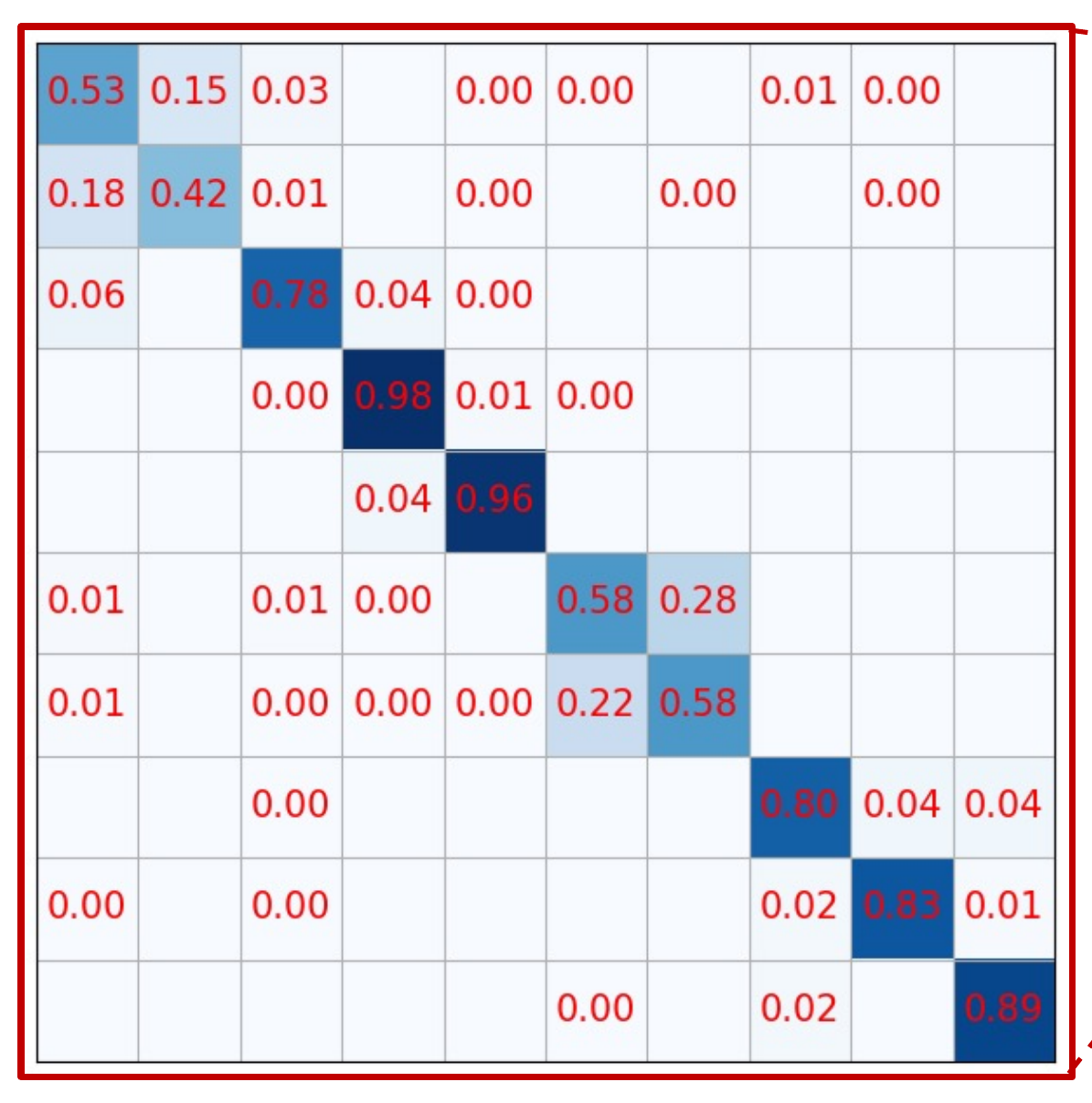}
  \label{NTUCS_cm}}
  \hfil
  \subfloat[]{\includegraphics[width=0.45\linewidth]{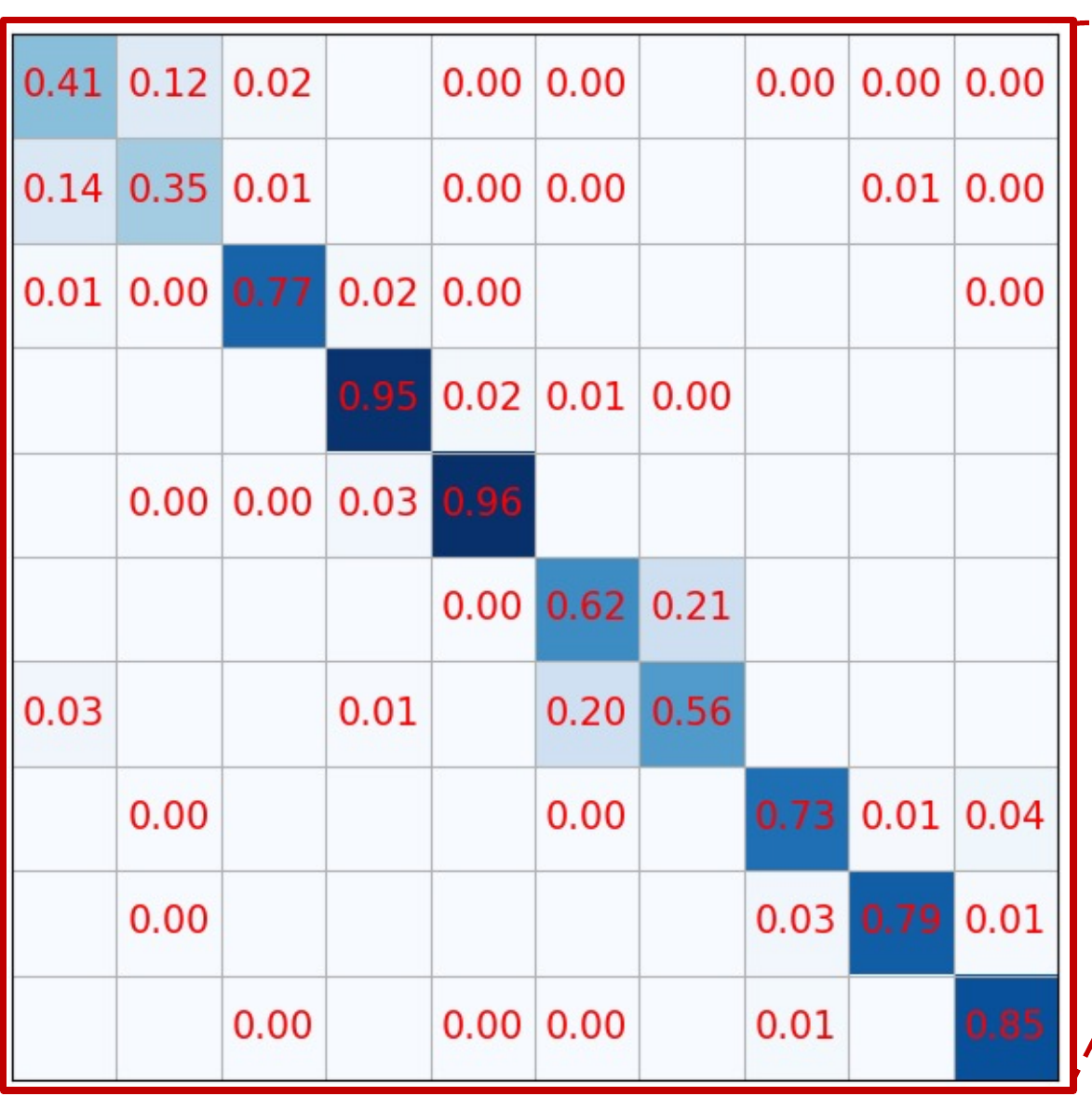}
  \label{NTU120CS_cm}}
  \hfil
  \subfloat[]{\includegraphics[width=0.45\linewidth]{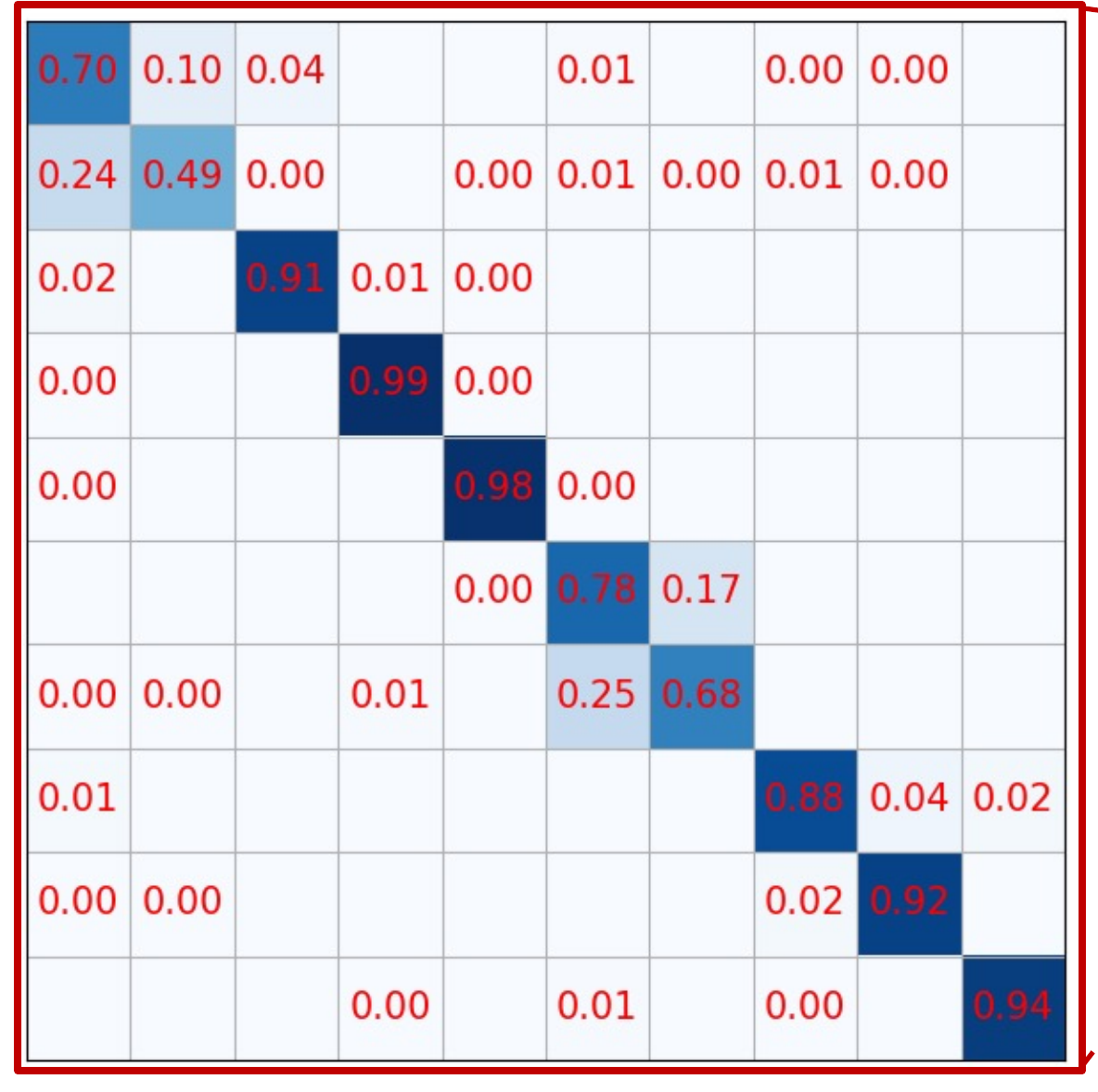}
  \label{NTUCV_cm}}
  \hfil
  \subfloat[]{\includegraphics[width=0.45\linewidth]{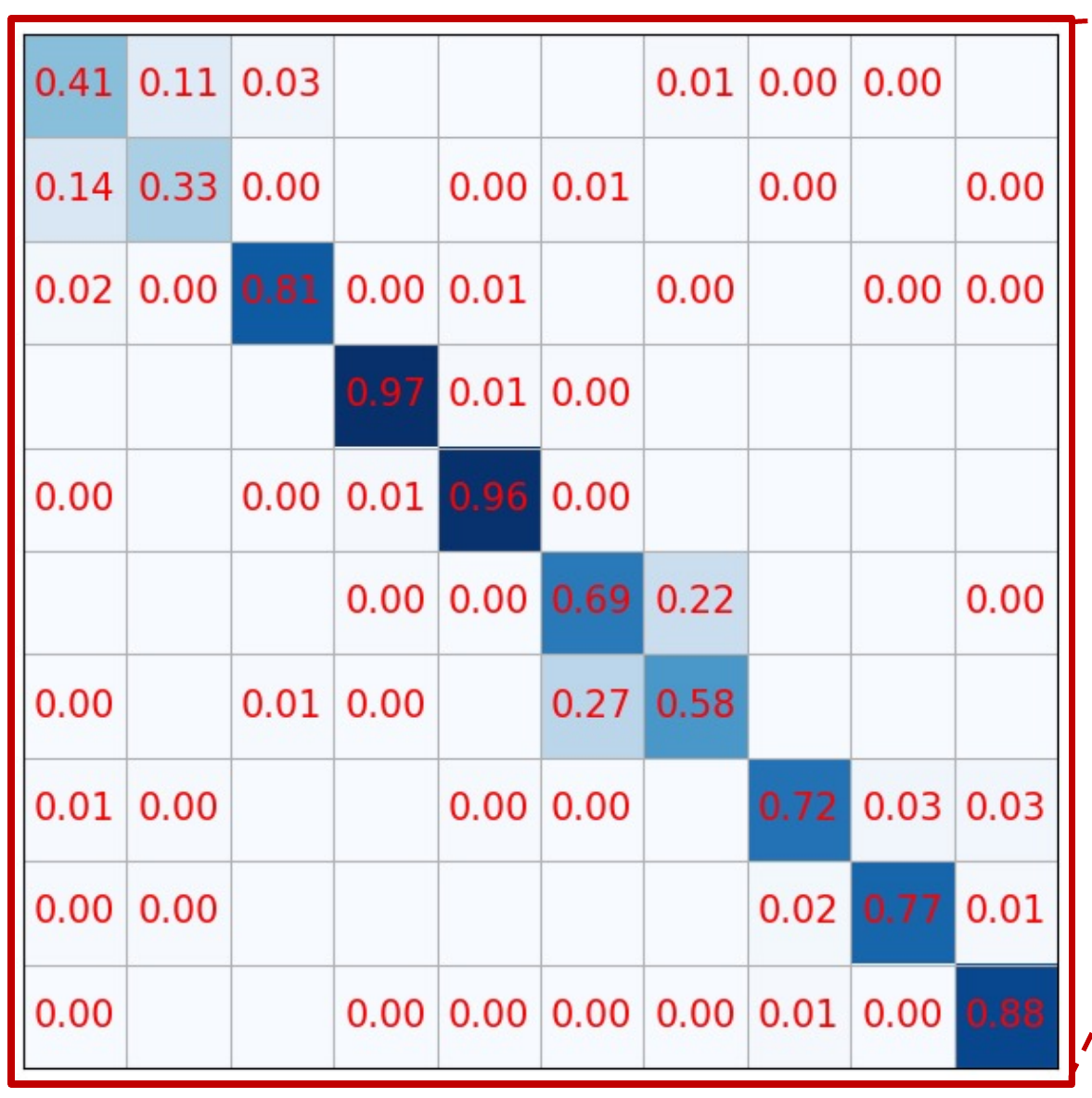}
  \label{NTU120CE_cm}}
  \caption{The confusion matrices obtained by A$^2$MC on NTU-60 and NTU-120 in the linear evaluation. The left confusion matrix of each group comes from classes 11 to 20. (a) On NTU-60 (x-sub). (b) On NTU-120 (x-sub). (c) On NTU-60 (x-view). (d) On NTU-120 (x-set).}
  \label{fig_cm}
\end{figure}

\noindent\textbf{Finetuned evaluation results on NTU-60.}
In the finetuned evaluation, the pre-trained query encoder and a linear classifier are trained on all labeled data in a fully-supervised way. Compare with the SOTA method (i.e., CPM~\cite{zhang2022contrastive}), the finetuned A$^2$MC achieves comparable performance, as shown in Table~\ref{finetuned}. This illustrates the generalization ability of A$^2$MC.

\noindent\textbf{Semi-supervised evaluation results on NTU-60.}
To verify the performance of ${\text A}^2$MC in the semi-supervised scenario, we conduct experiments on NTU-60 with a small amount of labeled data. As shown in Table~\ref{semi-supervised0}, ${\text A}^2$MC outperforms CMD~\cite{mao2022cmd} by 1.4\% and 1.3\% on x-sub with 1\% labeled data and x-view with 10\% labeled data, respectively.
In addition, we further provide the experimental results on randomly sampled 5\% and 20\% labeled data. As shown in Table~\ref{semi-supervised}, A$^2$MC performs well with different proportions of labeled data, proving the effectiveness of A$^2$MC.

\noindent\textbf{Linear evaluation results in three modalities on remain benchmarks.} In addition to the linear evaluation comparison results of three modalities on NTU-60 xsub, we further provide results on the remaining benchmarks, including NTU-60 xview and NTU-120 xsub/xset, as shown in Table~\ref{thrmodality}. Compared with SOTA methods (i.e.,, 3s-HiCo~\cite{dong2022hierarchical} and 3s-CMD~\cite{mao2022cmd}), the proposed 3s-A$^2$MC has comparable performance, proving its effectiveness.

\noindent\textbf{The change curves of loss and accuracy.}
Figure~\ref{fig_pre_curve} illustrates the changes of MC Loss, $\mathcal{L}_1$ loss (from weak attack-augment), $\mathcal{L}_2$ loss (from strong attack-augment), and $\mathcal{L}_3$ loss (from basic augment) in the pre-training contrastive learning. We can observe that: 1) the convergence rate of $\mathcal{L}_2$ is slower than that of $\mathcal{L}_1$ and $\mathcal{L}_3$; 2) all Losses further decrease at epoch 350 due to the drop of the learning rate, and finally stabilize at the epoch 750 on NTU-60 and NTU-120. Moreover, Figure~\ref{fig_lin_curve} shows the changes of recognition loss and accuracy in the linear evaluation process. We can find that: 1) the values of recognition loss and accuracy become steady at epoch 80; 2) decreasing the learning rate at epoch 50 is important for improving the recognition performance on NTU-60 and NTU-120.

\noindent\textbf{The confusion matrices obtained by A$^2$MC.}
As shown in Figure~\ref{fig_cm}, there are four groups of confusion matrices on NTU-60 (x-sub/x-view) and NTU-120 (x-sub/x-set) in the linear evaluation. In particular, the left confusion matrix of each group comes from classes 11 to 20. As can be seen from the Figure~\ref{fig_cm}: 1) the classification accuracy on NTU-60/120 (x-sub) is lower than that on NTU-60/120 (x-view/x-set), indicating that the difference in different subjects performing the same action is more challenging for A$^2$MC; 2) the classification accuracy of fine-grained similar actions (e.g., ``reading" vs ``writing" and ``wear a shoe" vs ``take off a shoe") on NTU-60 and NTU-120 is relatively low.

\begin{figure*}
\begin{center}
\includegraphics[width=0.9\linewidth]{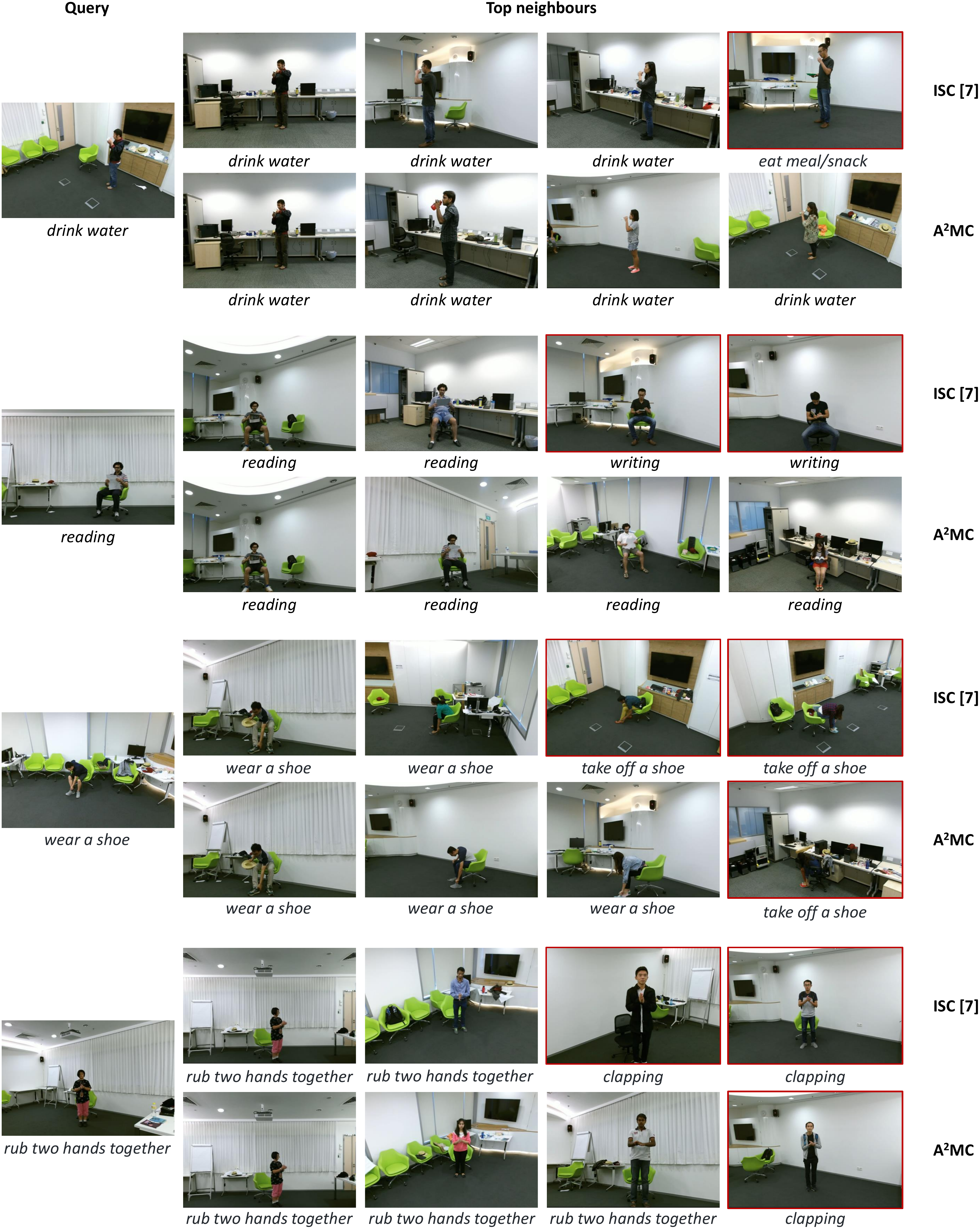}
\end{center}
   \caption{KNN evaluation results on NTU-60. Each group includes the query, and its Top-4 nearest neighbors produced by ISC~\cite{thoker2021skeleton} and A$^2$MC, respectively. All experimental results originally come from skeleton data, but we visualize the corresponding RGB videos instead of skeleton data for easy understanding.}
\label{fig:knn}
\end{figure*}

\noindent\textbf{Visualization of KNN evaluation results.} KNN evaluation results in Section 4.3 prove that A$^2$MC learns a high-quality feature space. Following ISC~\cite{thoker2021skeleton}, we also visualize the Top-4 nearest neighbors for the given query. As shown in Figure~\ref{fig:knn}, each group includes the query, and its Top-4 nearest neighbors produced by ISC~\cite{thoker2021skeleton} and A$^2$MC, respectively. All experimental results originally come from skeleton data, but we visualize the corresponding RGB videos instead of skeleton data for easy understanding. Compared with the results of ISC, the Top-4 neighbors learned by A$^2$MC contain more actions consistent with the query category. This indicates that the representations learned by the A$^2$MC are more discriminative for similar actions.

\section{Discussion}
{\textit{``A villain is an angel"}}. Attack is an angel we need in this work. The designed Attack-Augmentation (Att-Aug) is the first attempt to construct hard positive features by combining the advantages of attack and augmentation. The former perturbs the skeleton sequence mainly at the semantic level with the purpose of obtaining the features being close to the semantic boundary, while the latter perturbs the skeleton sequence mainly at the appearance level for randomly obtaining the diverse features. Specifically, to achieve such attacked result, we adopt the attack loss in an unsupervised way to smooth the feature distribution, in which the difference of top-k bars is suppressed to some extent. This is a simple yet effective way, even though there are various attack strategies that can also realize this purpose.

\section{Conclusion}
We propose a novel Attack-Augmentation Mixing-Contrastive skeletal representation learning (${\text A}^2$MC) framework for unsupervised skeleton-based action recognition. It integrates Attack-Augmentation (Att-Aug) mechanism and Positive-Negative Mixer (PNM) to generate hard positive/negative features for Mixing Contrast (MC). In particular, Att-Aug collaboratively implements targeted semantic attack and untargeted appearance augmentation of skeletons to facilitate the model to learn more robust representations. Extensive experiments demonstrate the superiority of ${\text A}^2$MC. As the first attempt, our attack in ${\text A}^2$MC is relatively straightforward, unlike the augmentation family within various weak/strong versions. This inspires us to further explore more attack-combination strategies in the future.




 
\bibliographystyle{IEEEtran}
\bibliography{IEEEabrv, main}

\begin{thebibliography}{10}
\providecommand{\url}[1]{#1}
\csname url@samestyle\endcsname
\providecommand{\newblock}{\relax}
\providecommand{\bibinfo}[2]{#2}
\providecommand{\BIBentrySTDinterwordspacing}{\spaceskip=0pt\relax}
\providecommand{\BIBentryALTinterwordstretchfactor}{4}
\providecommand{\BIBentryALTinterwordspacing}{\spaceskip=\fontdimen2\font plus
\BIBentryALTinterwordstretchfactor\fontdimen3\font minus \fontdimen4\font\relax}
\providecommand{\BIBforeignlanguage}[2]{{%
\expandafter\ifx\csname l@#1\endcsname\relax
\typeout{** WARNING: IEEEtran.bst: No hyphenation pattern has been}%
\typeout{** loaded for the language `#1'. Using the pattern for}%
\typeout{** the default language instead.}%
\else
\language=\csname l@#1\endcsname
\fi
#2}}
\providecommand{\BIBdecl}{\relax}
\BIBdecl

\bibitem{carreira2017quo}
J.~Carreira and A.~Zisserman, ``Quo vadis, action recognition? a new model and the kinetics dataset,'' in \emph{CVPR}, 2017.

\bibitem{shu2019hierarchical}
X.~Shu, J.~Tang, G.~Qi, W.~Liu, and J.~Yang, ``Hierarchical long short-term concurrent memory for human interaction recognition,'' \emph{IEEE TPAMI}, vol.~40, no.~3, pp. 1110--1118, 2021.

\bibitem{shu2020host}
X.~Shu, L.~Zhang, Y.~Sun, and J.~Tang, ``Host--parasite: Graph lstm-in-lstm for group activity recognition,'' \emph{IEEE TNNLS}, vol.~32, no.~2, pp. 663--674, 2021.

\bibitem{kumawat2022action}
S.~Kumawat, T.~Okawara, M.~Yoshida, H.~Nagahara, and Y.~Yagi, ``Action recognition from a single coded image,'' \emph{IEEE TPAMI}, vol.~45, no.~4, pp. 4109--4121, 2022.

\bibitem{li2022egocentric}
H.~Li, W.-S. Zheng, J.~Zhang, H.~Hu, J.~Lu, and J.-H. Lai, ``Egocentric action recognition by automatic relation modeling,'' \emph{IEEE TPAMI}, vol.~45, no.~1, pp. 489--507, 2022.

\bibitem{kim2017interpretable}
T.~S. Kim and A.~Reiter, ``Interpretable 3d human action analysis with temporal convolutional networks,'' in \emph{CVPR Workshops}, 2017.

\bibitem{li2018independently}
S.~Li, W.~Li, C.~Cook, C.~Zhu, and Y.~Gao, ``Independently recurrent neural network (indrnn): Building a longer and deeper rnn,'' in \emph{CVPR}, 2018.

\bibitem{shu2021spatiotemporal}
X.~Shu, L.~Zhang, G.-J. Qi, W.~Liu, and J.~Tang, ``Spatiotemporal co-attention recurrent neural networks for human-skeleton motion prediction,'' \emph{IEEE TPAMI}, vol.~44, no.~6, pp. 3300--3315, 2021.

\bibitem{shu2022multi}
X.~Shu, B.~Xu, L.~Zhang, and J.~Tang, ``Multi-granularity anchor-contrastive representation learning for semi-supervised skeleton-based action recognition,'' \emph{IEEE TPAMI}, vol.~45, no.~6, pp. 7559--7576, 2022.

\bibitem{li2019actional}
M.~Li, S.~Chen, X.~Chen, Y.~Zhang, Y.~Wang, and Q.~Tian, ``Actional-structural graph convolutional networks for skeleton-based action recognition,'' in \emph{CVPR}, 2019.

\bibitem{wang2021iip}
Q.~Wang, J.~Peng, S.~Shi, T.~Liu, J.~He, and R.~Weng, ``Iip-transformer: Intra-inter-part transformer for skeleton-based action recognition,'' \emph{arXiv}, 2021.

\bibitem{yan2018spatial}
S.~Yan, Y.~Xiong, and D.~Lin, ``Spatial temporal graph convolutional networks for skeleton-based action recognition,'' in \emph{AAAI}, 2018.

\bibitem{su2020predict}
K.~Su, X.~Liu, and E.~Shlizerman, ``Predict \& cluster: Unsupervised skeleton based action recognition,'' in \emph{CVPR}, 2020.

\bibitem{zheng2018unsupervised}
N.~Zheng, J.~Wen, R.~Liu, L.~Long, J.~Dai, and Z.~Gong, ``Unsupervised representation learning with long-term dynamics for skeleton based action recognition,'' in \emph{AAAI}, 2018.

\bibitem{guo2022contrastive}
T.~Guo, H.~Liu, Z.~Chen, M.~Liu, T.~Wang, and R.~Ding, ``Contrastive learning from extremely augmented skeleton sequences for self-supervised action recognition,'' in \emph{AAAI}, 2022.

\bibitem{li20213d}
L.~Li, M.~Wang, B.~Ni, H.~Wang, J.~Yang, and W.~Zhang, ``3d human action representation learning via cross-view consistency pursuit,'' in \emph{CVPR}, 2021.

\bibitem{cheng2021hierarchical}
Y.-B. Cheng, X.~Chen, J.~Chen, P.~Wei, D.~Zhang, and L.~Lin, ``Hierarchical transformer: Unsupervised representation learning for skeleton-based human action recognition,'' in \emph{ICME}, 2021.

\bibitem{kim2022global}
B.~Kim, H.~J. Chang, J.~Kim, and J.~Y. Choi, ``Global-local motion transformer for unsupervised skeleton-based action learning,'' in \emph{ECCV}, 2022.

\bibitem{chen2022hierarchically}
Y.~Chen, L.~Zhao, J.~Yuan, Y.~Tian, Z.~Xia, S.~Geng, L.~Han, and D.~N. Metaxas, ``Hierarchically self-supervised transformer for human skeleton representation learning,'' in \emph{ECCV}, 2022.

\bibitem{chen2020simple}
T.~Chen, S.~Kornblith, M.~Norouzi, and G.~Hinton, ``A simple framework for contrastive learning of visual representations,'' in \emph{ICML}, 2020.

\bibitem{he2020momentum}
K.~He, H.~Fan, Y.~Wu, S.~Xie, and R.~Girshick, ``Momentum contrast for unsupervised visual representation learning,'' in \emph{CVPR}, 2020.

\bibitem{gutmann2010noise}
M.~Gutmann and A.~Hyv{\"a}rinen, ``Noise-contrastive estimation: A new estimation principle for unnormalized statistical models,'' in \emph{AISTATS}, 2010.

\bibitem{robinson2020contrastive}
J.~Robinson, C.-Y. Chuang, S.~Sra, and S.~Jegelka, ``Contrastive learning with hard negative samples,'' \emph{arXiv}, 2020.

\bibitem{lin2020ms2l}
L.~Lin, S.~Song, W.~Yang, and J.~Liu, ``Ms2l: Multi-task self-supervised learning for skeleton based action recognition,'' in \emph{ACM MM}, 2020.

\bibitem{kalantidis2020hard}
Y.~Kalantidis, M.~B. Sariyildiz, N.~Pion, P.~Weinzaepfel, and D.~Larlus, ``Hard negative mixing for contrastive learning,'' in \emph{NeurIPS}, 2020.

\bibitem{liu2020adversarial}
J.~Liu, N.~Akhtar, and A.~Mian, ``Adversarial attack on skeleton-based human action recognition,'' \emph{IEEE TNNLS}, vol.~33, no.~4, pp. 1609--1622, 2020.

\bibitem{wang2021understanding}
H.~Wang, F.~He, Z.~Peng, T.~Shao, Y.-L. Yang, K.~Zhou, and D.~Hogg, ``Understanding the robustness of skeleton-based action recognition under adversarial attack,'' in \emph{CVPR}, 2021.

\bibitem{li2020sparse}
J.~Li and E.~Shlizerman, ``Sparse semi-supervised action recognition with active learning,'' \emph{arXiv}, 2020.

\bibitem{mao2022cmd}
Y.~Mao, W.~Zhou, Z.~Lu, J.~Deng, and H.~Li, ``Cmd: Self-supervised 3d action representation learning with cross-modal mutual distillation,'' in \emph{ECCV}, 2022.

\bibitem{si2020adversarial}
C.~Si, X.~Nie, W.~Wang, L.~Wang, T.~Tan, and J.~Feng, ``Adversarial self-supervised learning for semi-supervised 3d action recognition,'' in \emph{ECCV}, 2020.

\bibitem{su2021modeling}
Y.~Su, G.~Lin, R.~Sun, Y.~Hao, and Q.~Wu, ``Modeling the uncertainty for self-supervised 3d skeleton action representation learning,'' in \emph{ACM MM}, 2021.

\bibitem{su2021self}
Y.~Su, G.~Lin, and Q.~Wu, ``Self-supervised 3d skeleton action representation learning with motion consistency and continuity,'' in \emph{ICCV}, 2021.

\bibitem{thoker2021skeleton}
F.~M. Thoker, H.~Doughty, and C.~G. Snoek, ``Skeleton-contrastive 3d action representation learning,'' in \emph{ACM MM}, 2021.

\bibitem{tu2022joint}
Z.~Tu, J.~Zhang, H.~Li, Y.~Chen, and J.~Yuan, ``Joint-bone fusion graph convolutional network for semi-supervised skeleton action recognition,'' \emph{IEEE TMM}, 2022.

\bibitem{xu2021prototypical}
S.~Xu, H.~Rao, X.~Hu, J.~Cheng, and B.~Hu, ``Prototypical contrast and reverse prediction: Unsupervised skeleton based action recognition,'' \emph{IEEE TMM}, vol.~25, pp. 624--634, 2021.

\bibitem{xu2021unsupervised}
Z.~Xu, X.~Shen, Y.~Wong, and M.~S. Kankanhalli, ``Unsupervised motion representation learning with capsule autoencoders,'' in \emph{NeurIPS}, 2021.

\bibitem{yang2021skeleton}
S.~Yang, J.~Liu, S.~Lu, M.~H. Er, and A.~C. Kot, ``Skeleton cloud colorization for unsupervised 3d action representation learning,'' in \emph{ICCV}, 2021.

\bibitem{yao2021recurrent}
H.~Yao, S.-J. Zhao, C.~Xie, K.~Ye, and S.~Liang, ``Recurrent graph convolutional autoencoder for unsupervised skeleton-based action recognition,'' in \emph{ICME}, 2021.

\bibitem{zhang2022contrastive}
H.~Zhang, Y.~Hou, W.~Zhang, and W.~Li, ``Contrastive positive mining for unsupervised 3d action representation learning,'' in \emph{ECCV}, 2022.

\bibitem{mao2023masked}
Y.~Mao, J.~Deng, W.~Zhou, Y.~Fang, W.~Ouyang, and H.~Li, ``Masked motion predictors are strong 3d action representation learners,'' in \emph{ICCV}, 2023, pp. 10\,181--10\,191.

\bibitem{chen2020improved}
X.~Chen, H.~Fan, R.~Girshick, and K.~He, ``Improved baselines with momentum contrastive learning,'' \emph{arXiv}, 2020.

\bibitem{lin2023actionlet}
L.~Lin, J.~Zhang, and J.~Liu, ``Actionlet-dependent contrastive learning for unsupervised skeleton-based action recognition,'' in \emph{CVPR}, 2023, pp. 2363--2372.

\bibitem{shah2023halp}
A.~Shah, A.~Roy, K.~Shah, S.~Mishra, D.~Jacobs, A.~Cherian, and R.~Chellappa, ``Halp: Hallucinating latent positives for skeleton-based self-supervised learning of actions,'' in \emph{CVPR}, 2023, pp. 18\,846--18\,856.

\bibitem{rao2021augmented}
H.~Rao, S.~Xu, X.~Hu, J.~Cheng, and B.~Hu, ``Augmented skeleton based contrastive action learning with momentum lstm for unsupervised action recognition,'' \emph{Information Sciences}, vol. 569, pp. 90--109, 2021.

\bibitem{zhang2022hierarchical}
J.~Zhang, L.~Lin, and J.~Liu, ``Hierarchical consistent contrastive learning for skeleton-based action recognition with growing augmentations,'' \emph{arXiv}, 2022.

\bibitem{szegedy2013intriguing}
C.~Szegedy, W.~Zaremba, I.~Sutskever, J.~Bruna, D.~Erhan, I.~Goodfellow, and R.~Fergus, ``Intriguing properties of neural networks,'' \emph{arXiv}, 2013.

\bibitem{tanaka2022adversarial}
N.~Tanaka, H.~Kera, and K.~Kawamoto, ``Adversarial bone length attack on action recognition,'' in \emph{AAAI}, 2022, pp. 2335--2343.

\bibitem{zhang2024meta}
L.~Zhang, Y.~Zhou, Y.~Yang, and X.~Gao, ``Meta invariance defense towards generalizable robustness to unknown adversarial attacks,'' \emph{IEEE TPAMI}, 2024.

\bibitem{akhtar2021advances}
N.~Akhtar, A.~Mian, N.~Kardan, and M.~Shah, ``Advances in adversarial attacks and defenses in computer vision: A survey,'' \emph{IEEE Access}, vol.~9, pp. 155\,161--155\,196, 2021.

\bibitem{goodfellow2014explaining}
I.~J. Goodfellow, J.~Shlens, and C.~Szegedy, ``Explaining and harnessing adversarial examples,'' \emph{arXiv}, 2014.

\bibitem{madry2017towards}
A.~Madry, A.~Makelov, L.~Schmidt, D.~Tsipras, and A.~Vladu, ``Towards deep learning models resistant to adversarial attacks,'' \emph{arXiv}, 2017.

\bibitem{oord2018representation}
A.~v.~d. Oord, Y.~Li, and O.~Vinyals, ``Representation learning with contrastive predictive coding,'' \emph{arXiv}, 2018.

\bibitem{kingma2014adam}
D.~P. Kingma and J.~Ba, ``Adam: A method for stochastic optimization,'' \emph{arXiv}, 2014.

\bibitem{hu2021adco}
Q.~Hu, X.~Wang, W.~Hu, and G.-J. Qi, ``Adco: Adversarial contrast for efficient learning of unsupervised representations from self-trained negative adversaries,'' in \emph{CVPR}, 2021.

\bibitem{qi2022adversarial}
G.-J. Qi and M.~Shah, ``Adversarial pretraining of self-supervised deep networks: Past, present and future,'' \emph{arXiv}, 2022.

\bibitem{shahroudy2016ntu}
A.~Shahroudy, J.~Liu, T.-T. Ng, and G.~Wang, ``Ntu rgb+ d: A large scale dataset for 3d human activity analysis,'' in \emph{CVPR}, 2016.

\bibitem{liu2019ntu}
J.~Liu, A.~Shahroudy, M.~Perez, G.~Wang, L.-Y. Duan, and A.~C. Kot, ``Ntu rgb+ d 120: A large-scale benchmark for 3d human activity understanding,'' \emph{IEEE TPAMI}, vol.~42, no.~10, pp. 2684--2701, 2019.

\bibitem{liu2017pku}
C.~Liu, Y.~Hu, Y.~Li, S.~Song, and J.~Liu, ``Pku-mmd: A large scale benchmark for continuous multi-modal human action understanding,'' \emph{arXiv}, 2017.

\bibitem{shi2020decoupled}
L.~Shi, Y.~Zhang, J.~Cheng, and H.~Lu, ``Decoupled spatial-temporal attention network for skeleton-based action-gesture recognition,'' in \emph{ACCV}, 2020.

\bibitem{wu2024scd}
C.~Wu, X.-J. Wu, J.~Kittler, T.~Xu, S.~Ahmed, M.~Awais, and Z.~Feng, ``Scd-net: Spatiotemporal clues disentanglement network for self-supervised skeleton-based action recognition,'' in \emph{AAAI}, 2024, pp. 5949--5957.

\bibitem{paszke2017automatic}
A.~Paszke, S.~Gross, S.~Chintala, G.~Chanan, E.~Yang, Z.~DeVito, Z.~Lin, A.~Desmaison, L.~Antiga, and A.~Lerer, ``Automatic differentiation in pytorch,'' in \emph{NeurIPS Workshops}, 2017.

\bibitem{kundu2019unsupervised}
J.~N. Kundu, M.~Gor, P.~K. Uppala, and V.~B. Radhakrishnan, ``Unsupervised feature learning of human actions as trajectories in pose embedding manifold,'' in \emph{WACV}, 2019.

\bibitem{nie2020unsupervised}
Q.~Nie, Z.~Liu, and Y.~Liu, ``Unsupervised 3d human pose representation with viewpoint and pose disentanglement,'' in \emph{ECCV}, 2020.

\bibitem{wang2021contrast}
P.~Wang, J.~Wen, C.~Si, Y.~Qian, and L.~Wang, ``Contrast-reconstruction representation learning for self-supervised skeleton-based action recognition,'' \emph{IEEE TIP}, vol.~31, pp. 6224--6238, 2022.

\bibitem{zhu2023modeling}
Y.~Zhu, H.~Han, Z.~Yu, and G.~Liu, ``Modeling the relative visual tempo for self-supervised skeleton-based action recognition,'' in \emph{ICCV}, 2023, pp. 13\,913--13\,922.

\bibitem{zhou2023self}
Y.~Zhou, H.~Duan, A.~Rao, B.~Su, and J.~Wang, ``Self-supervised action representation learning from partial spatio-temporal skeleton sequences,'' in \emph{AAAI}, 2023, pp. 3825--3833.

\bibitem{yang2024view}
D.~Yang, Y.~Wang, A.~Dantcheva, L.~Garattoni, G.~Francesca, and F.~Br{\'e}mond, ``View-invariant skeleton action representation learning via motion retargeting,'' \emph{IJCV}, vol. 132, pp. 2351--2366, 2024.

\bibitem{francohyperbolic}
L.~Franco, P.~Mandica, B.~Munjal, and F.~Galasso, ``Hyperbolic self-paced learning for self-supervised skeleton-based action representations,'' in \emph{ICLR}, 2023.

\bibitem{hua2023part}
Y.~Hua, W.~Wu, C.~Zheng, A.~Lu, M.~Liu, C.~Chen, and S.~Wu, ``Part aware contrastive learning for self-supervised action recognition,'' in \emph{IJCAI}, 2023, pp. 855--863.

\bibitem{zhang2023hierarchical}
J.~Zhang, L.~Lin, and J.~Liu, ``Hierarchical consistent contrastive learning for skeleton-based action recognition with growing augmentations,'' in \emph{AAAI}, 2023, pp. 3427--3435.

\bibitem{sun2023unified}
S.~Sun, D.~Liu, J.~Dong, X.~Qu, J.~Gao, X.~Yang, X.~Wang, and M.~Wang, ``Unified multi-modal unsupervised representation learning for skeleton-based action understanding,'' in \emph{ACM MM}, 2023, pp. 2973--2984.

\bibitem{lin2024mutual}
L.~Lin, J.~Zhang, and J.~Liu, ``Mutual information driven equivariant contrastive learning for 3d action representation learning,'' \emph{IEEE TIP}, vol.~33, pp. 1883--1897, 2024.

\bibitem{zhang2023prompted}
J.~Zhang, L.~Lin, and J.~Liu, ``Prompted contrast with masked motion modeling: Towards versatile 3d action representation learning,'' in \emph{ACM MM}, 2023, pp. 7175--7183.

\bibitem{dong2022hierarchical}
J.~Dong, S.~Sun, Z.~Liu, S.~Chen, B.~Liu, and X.~Wang, ``Hierarchical contrast for unsupervised skeleton-based action representation learning,'' \emph{arXiv}, 2022.

\bibitem{chen2022contrastive}
Z.~Chen, H.~Liu, T.~Guo, Z.~Chen, P.~Song, and H.~Tang, ``Contrastive learning from spatio-temporal mixed skeleton sequences for self-supervised skeleton-based action recognition,'' \emph{arXiv}, 2022.

\end{thebibliography}

\begin{IEEEbiography}[{\includegraphics[width=1in,height=1.25in,clip,keepaspectratio]{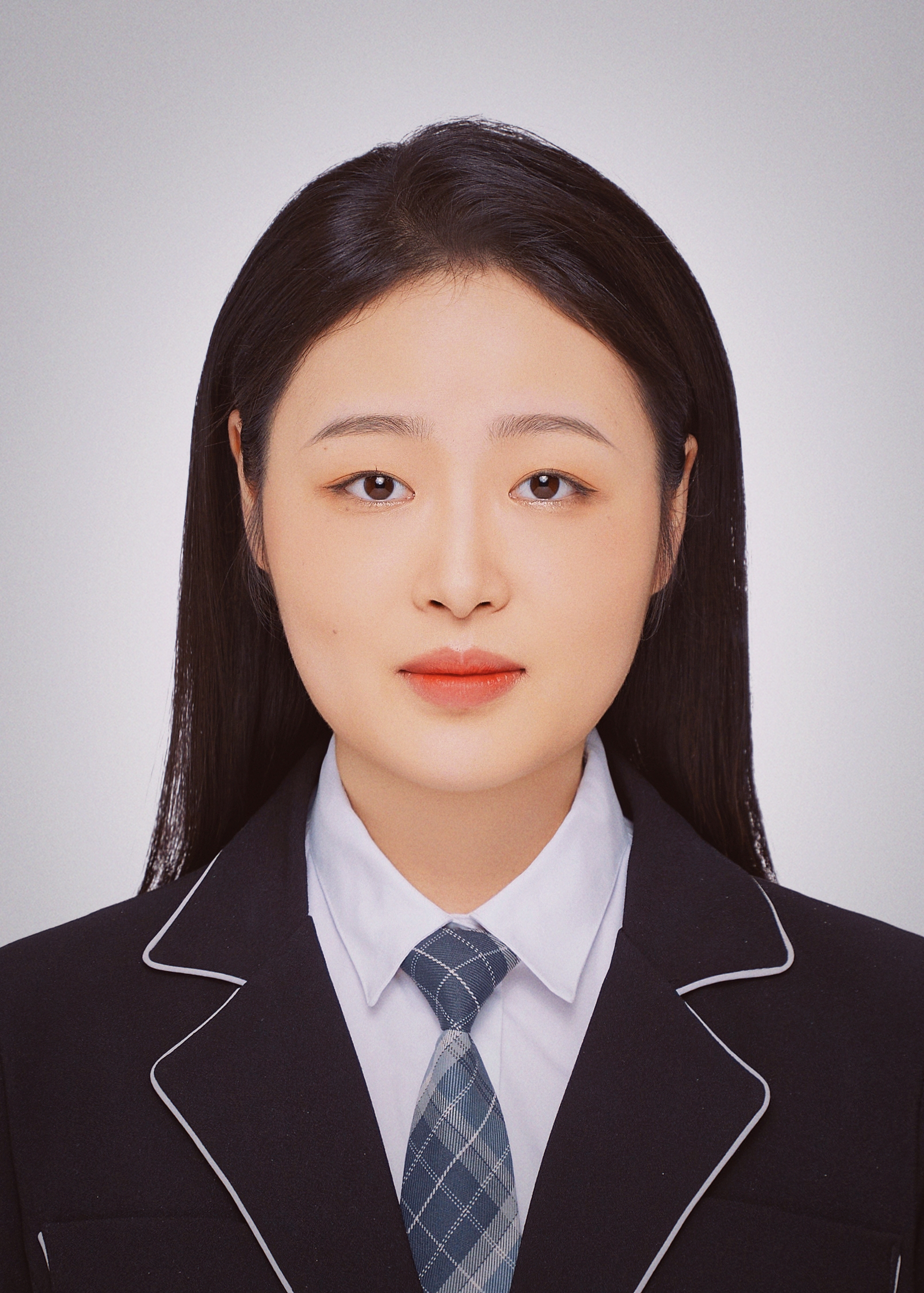}}]{Binqian~Xu}
is currently a Ph.D. Degree Candidate in the School of Computer Science and Engineering, Nanjing University of Science and Technology, China. Now, she is also as a visiting scholar at the National University of Singapore. Her current research interest is Computer Vision, and Deep Learning. She has authored several journal/conference papers in these areas, e.g., IEEE TPAMI, IEEE TIP, NeurIPS, etc.
\end{IEEEbiography}

\begin{IEEEbiography}[{\includegraphics[width=1in,height=1.25in,clip,keepaspectratio]{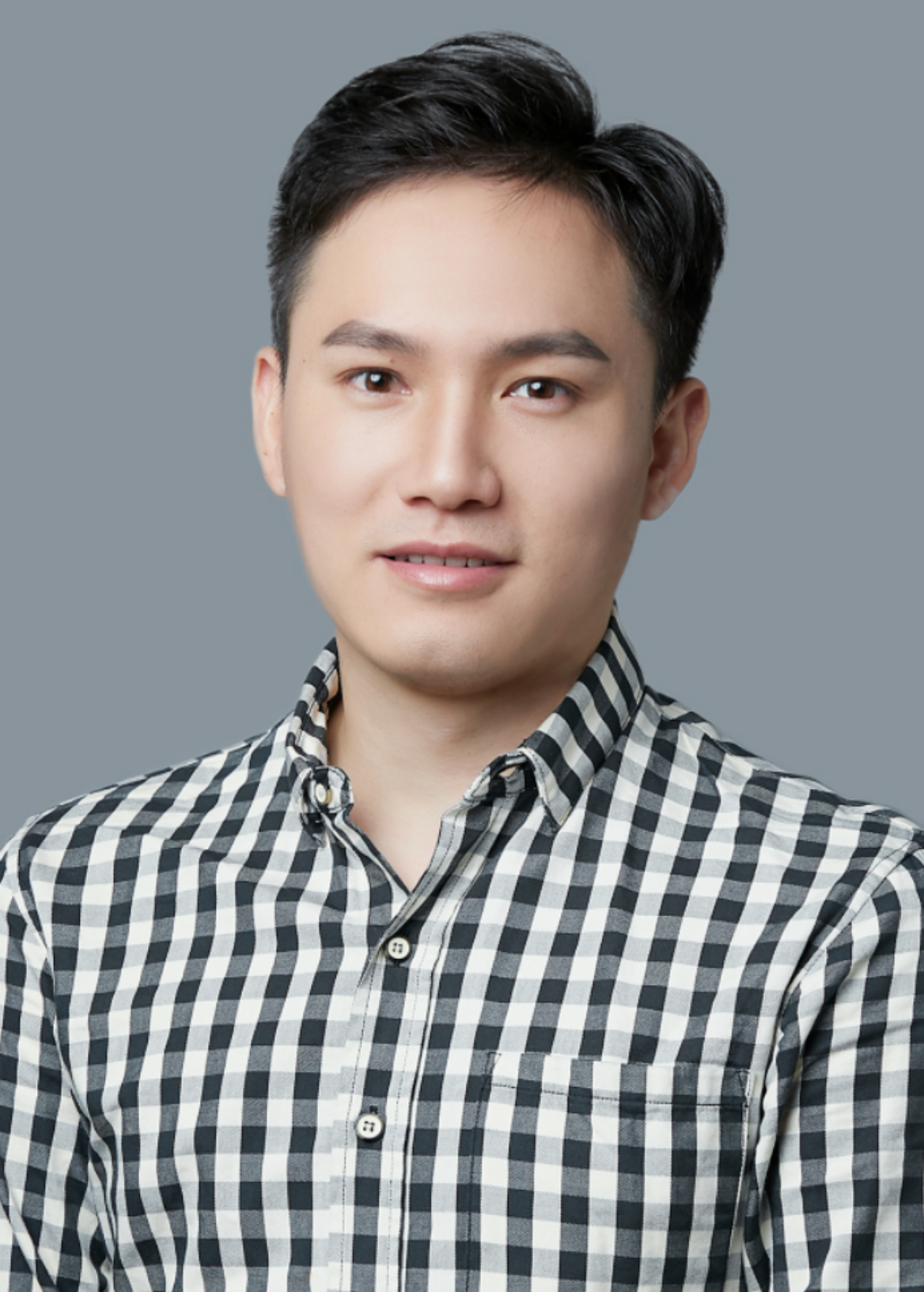}}]{Xiangbo~Shu}
 is currently a Professor in School of Computer Science and Engineering, Nanjing Univesity of Science and Technology, China. Before that, he also worked as a visiting scholar in National University of Singapore, Singapore. His current research interests include Computer Vision, and Multimedia. He has authored over 100 journal and conference papers in these areas, including IEEE TPAMI, IEEE TNNLS, IEEE TIP, CVPR, ICCV, ECCV, ACM MM, etc. He has received the Best Student Paper Award in MMM 2016, and the Best Paper Runner-up in ACM MM 2015. He has served as the editorial
boards of the IEEE TNNLs, IEEE TCSVT, Pattern Recognition, etc. He is also the Member of ACM, the Senior Member of CCF, and the Senior Member of IEEE.
\end{IEEEbiography}

\begin{IEEEbiography}[{\includegraphics[width=1in,height=1.25in,clip,keepaspectratio]{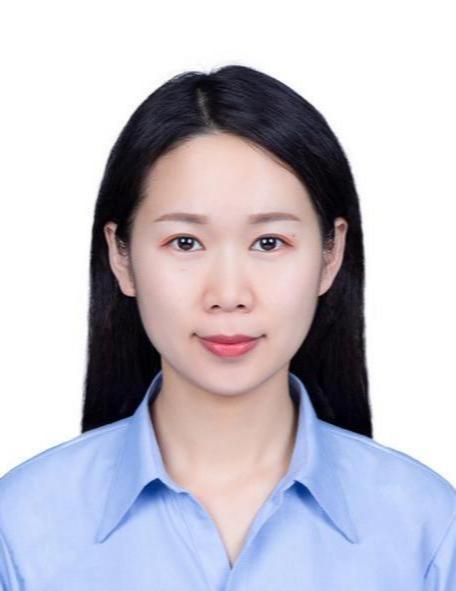}}]{Jiachao~Zhang}
is currently an Associate Professor in Artificial Intelligence Industrial Technology Research Institute, Nanjing Institute of Technology, China. She received her Bachelor's degree from Nanjing University of Science and Technology in 2011, Master's degree University of Dayton in 2015, and Ph.D. degree from Nanjing University of Science and Technology in 2018. She visited University of Dayton as a visiting scholar from 2016 to 2017. Her current research interest is Image Processing, and Computer Vision. She is the Member of CCF, and the Member of IEEE.
\end{IEEEbiography}

\begin{IEEEbiography}[{\includegraphics[width=1in,height=1.25in,clip,keepaspectratio]{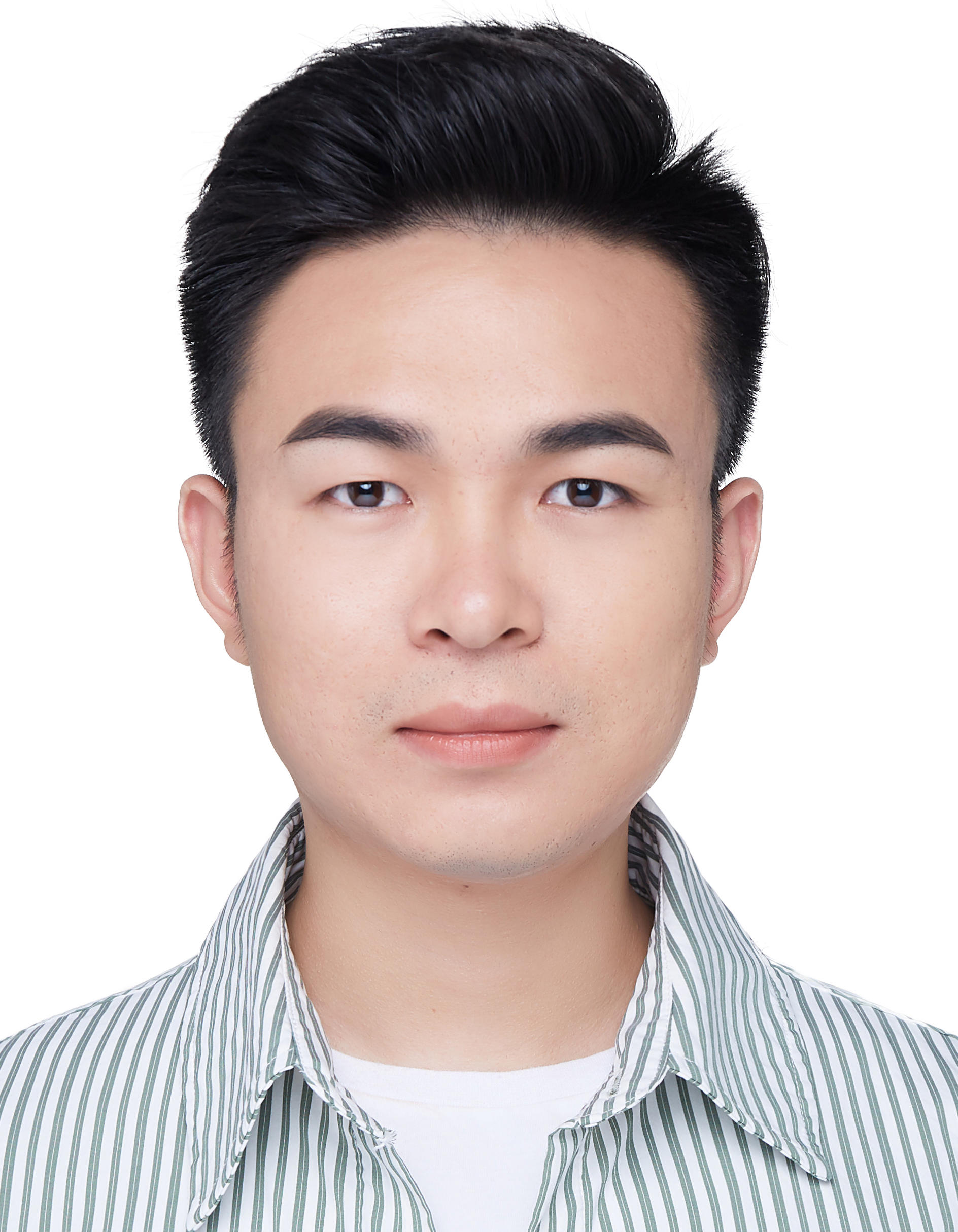}}]{Rui~Yan}
received the Ph.D. degree at Intelligent
Media Analysis Group (IMAG), Nanjing University
of Science and Technology, China. He is currently an Assistant Researcher at the Department of Computer Science and Technology, Nanjing University, China. He was a research intern (part-time) at ByteDance from Jan. 2022 to Aug. 2022. He was a research intern (part-time) at Tencent from Sep. 2021 to Dec. 2021. He was a visiting researcher at the National University of Singapore (NUS) from Aug. 2021 to Aug. 2022. He was a research intern at HUAWEI NOAH’S ARK LAB from Dec. 2018 to Dec. 2019. His research mainly focuses on Complex Human Behavior Understanding and Video-Language Understanding. He has authored over 20 journal and conference papers in these areas, including IEEE TPAMI, IEEE TNNLS, IEEE TCSVT, CVPR,
NeurIPS, ECCV, and ACM MM, etc.
\end{IEEEbiography}

\begin{IEEEbiography}[{\includegraphics[width=1in,height=1.25in,clip,keepaspectratio]{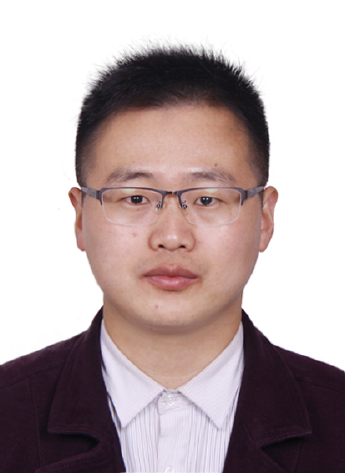}}]{Guo-Sen~Xie}
received the Ph.D. degree in pattern recognition and intelligent systems from
the National Laboratory of Pattern Recognition, Institute of Automation, Chinese Academy of
Sciences, Beijing, China, in 2016. From 2014 to 2015, he worked as a visiting scholar at the National University of Singapore, Singapore. He is a Professor at the School of Computer Science and Engineering, Nanjing University of Science and Technology, China. His research results have been expounded in more than 30 publications at prestigious journals and prominent conferences, such
as IEEE TPAMI, IJCV, IEEE TIP, IEEE TNNLS, NeurIPS, CVPR, ICCV, and ECCV. He has received the Best Student Paper Award in MMM 2016. His research interests include computer vision and machine learning.
\end{IEEEbiography}

\vfill

\end{document}